%% file: main.tex
\pdfoutput=1
\documentclass[11pt]{article}
\usepackage[]{acl}

\usepackage{times}
\usepackage{latexsym}
\usepackage{diagbox}
\usepackage{amssymb}

\usepackage[T1]{fontenc}
\usepackage[utf8]{inputenc}

\usepackage{microtype}
\usepackage{inconsolata}
\usepackage{cleveref}
\usepackage{placeins}
\usepackage{enumitem}
\usepackage{booktabs}
\usepackage{multicol}
\usepackage{multirow}
\usepackage{makecell}
\usepackage{arydshln}
\usepackage{tabularx}
\usepackage{CJKutf8}

\usepackage{graphicx}
\usepackage{dblfloatfix}
\usepackage{float}

\usepackage[font=small]{caption}
\usepackage[font=small]{subcaption}

\usepackage{tikz}
\usetikzlibrary{shapes}
\usepackage{pgfplots}
\usepackage{pdflscape}
\usepackage{xcolor}

\pgfplotsset{
    compat=1.17,
    xtick pos=bottom,
    ytick pos=left,
    every non boxed x axis/.append style={x axis line style=-},
    every non boxed y axis/.append style={y axis line style=-},
    every node near coord/.style={font=\tiny},
    scaled y ticks=false,
    area legend/.style={
        legend image code/.code={
            \draw[#1](0cm,-0.1cm)rectangle
                (1.5ex,0.08cm)
            ;
        }
    },
}

\title{Monolingual or Multilingual Instruction Tuning: \\ Which Makes a Better Alpaca}

\author{
Pinzhen Chen\textsuperscript{1,*}\qquad 
Shaoxiong Ji\textsuperscript{2,*}\qquad  
Nikolay Bogoychev\textsuperscript{1}\\ 
{\bf
Andrey Kutuzov\textsuperscript{3}\qquad
Barry Haddow\textsuperscript{1}\qquad  
Kenneth Heafield\textsuperscript{1}
}\\
\textsuperscript{1}University of Edinburgh \qquad
\textsuperscript{2}University of Helsinki \qquad
\textsuperscript{3}University of Oslo 
\\
\texttt{pchen3@ed.ac.uk\qquad shaoxiong.ji@helsinki.fi}}

\begin{document}
\maketitle
\def\thefootnote{*}\footnotetext{Equal contribution. Our code, training data, and test data will be at \url{https://github.com/hplt-project/monolingual-multilingual-instruction-tuning}.}
\def\thefootnote{\arabic{footnote}}

\begin{abstract}
Foundational large language models (LLMs) can be instruction-tuned to perform open-domain question answering, facilitating applications like chat assistants. While such efforts are often carried out in a single language, we empirically analyze cost-efficient strategies for multilingual scenarios. Our study employs the Alpaca dataset and machine translations of it to form multilingual data, which is then used to tune LLMs through either low-rank adaptation or full-parameter training. Under a controlled computation budget, comparisons show that multilingual tuning is on par or better than tuning a model for each language. Furthermore, multilingual tuning with downsampled data can be as powerful and more robust. Our findings serve as a guide for expanding language support through instruction tuning.
\end{abstract}

\section{Introduction}
Language capacity has attracted much attention in pre-trained language models.
Some pioneering works focused on a single language \citep{peters-etal-2018-deep,devlin-etal-2019-bert}, while later works aim to cover multiple languages \citep{conneau-etal-2020-unsupervised,liu-etal-2020-multilingual-denoising}. In the recent blossom of open-source LLMs, English-centric ones include GPT-2, LLaMA, and Pythia \citep{radford2019language,touvron2023llama,biderman2023pythia}, and multilingual ones are represented by BLOOM \citep{scao2022bloom}. Multilingual models seem attractive when considering operational costs, cross-lingual transfer, and low-resource languages \citep{artetxe-schwenk-2019-massively,wu-dredze-2020-languages}, yet English-centric models can possess good multilingual transferability \citep{ye2023language}.

Instruction tuning makes LLMs follow and respond to inputs \citep{sanh2021multitask,wei2022finetuned}. With multilingual instruction data becoming feasible and available, this paper compares monolingual and multilingual instruction tuning applied to English-centric and multilingual LLMs to search for the optimal strategy to support multiple languages. Unlike prior works on multilingual multi-NLP-task tuning \citep{mishra-etal-2022-cross,muennighoff-etal-2023-crosslingual-xp3}, we focus on open-ended question answering under language generation.

Our data setting combines two low-cost practices: self-instruct, which distils data from a powerful LLM \cite{wang-etal-2023-self-instruct,alpaca} and the idea of leveraging machine translation to create multilingual datasets \cite{muennighoff-etal-2023-crosslingual-xp3}. We fine-tune several decoder LLMs with either full-parameter fine-tuning (FFT) or low-rank adaptation \citep[LoRA,][]{hu2022lora} with different language combinations. Our experiments feature a fixed computation budget to offer practical insights. It is shown that multilingual tuning is preferred to monolingual tuning for each language under LoRA, but the results are mixed under FFT. English-tuned LLMs are not well-versed in responding in other languages, whereas a downsampled multilingual tuning scheme proposed by us is more robust. Finally, we examine our model performance on unseen languages and various LLMs of roughly the same size.

\section{Methodology}
\subsection{Instruction data}

We use the Alpaca dataset as a seed to create a multilingual instruction-response dataset. We used the cleaned version with 52K instances\footnote{\url{https://github.com/gururise/alpacadatacleaned}} and machine-translated it into eight languages: Bulgarian, Czech, Chinese, German, Finnish, French, Russian, and Spanish, using open-source translation systems.\footnote{\url{https://github.com/browsermt/bergamot-translator}}

\subsection{Budget-controlled instruction tuning}

For monolingual tuning, we tune LLMs for each language separately, whereas for multilingual tuning, we merge and shuffle the data in all languages. This allows for resource-controlled comparisons between monolingual and multilingual tuning, where a fixed (and equal for each language) computation budget is allocated to support all languages of interest. Experimental resource usage is described as follows:
\begin{itemize}[nolistsep,noitemsep]
    \item [1)] Let \textit{C\textsubscript{Alpaca}} denote the cost of \emph{monolingual} Alpaca fine-tuning for a single language, then it costs \textit{N$\times$C\textsubscript{Alpaca}} to tune individual models to support $N$ languages.
    \item [2)] \emph{Multilingual} instruction tuning will cost \textit{N$\times$C\textsubscript{Alpaca}} too, as it trains on data available in all $N$ languages in one go.
\end{itemize}
We can fairly compare LLMs trained via 1) and 2) for any language. In addition, we propose to benchmark two budget-saving options which cost the same \textit{C\textsubscript{Alpaca}} as a monolingual Alpaca:
\begin{itemize}[nolistsep,noitemsep]
    \item [3)] As a simple baseline, we use an \emph{English-tuned} model to respond to all languages.
    \item [4)] \emph{Downsampled multilingual}: we randomly sample from the multilingual data in 2) to have the size of a monolingual dataset.
\end{itemize}

Our study covers two training paradigms: \emph{low-rank adaptation} and \emph{full-parameter fine-tuning}. Both fine-tune an LLM with the causal language modelling objective on the instruction-response data, with hyperparameters listed in \Cref{app:exp-setup}.
Five LLMs are involved: Baichuan-2, BLOOM, LLaMA, OpenLLaMA, and Pythia, aiming to test with different language coverage in the base LLMs. Pythia, LLaMA, and OpenLLaMA are predominantly English, while Baichuan-2 and BLOOM are more versatile. A detailed description of the LLMs is in \Cref{app:exp-base-model}.

\subsection{Evaluation setup}
\label{sec:evaluation}

\paragraph{Test data}
Our instruction-tuned LLMs are benchmarked on languages both \emph{seen}  and \emph{unseen} during fine-tuning. We employ native speakers to manually translate 50 prompts sampled from OpenAssistant \citep{openassistant} into eight languages: six seen during training and two unseen. The seen category includes English, French, Spanish, Bulgarian, Russian, and Chinese. Among the six, English is the highest-resourced, followed by French and Spanish which share the same script as English. Bulgarian and Russian are European languages but use a writing system distinct from English. Finally, Chinese is a high-resource distant language in a different script. For unseen tests, we pick Bengali and Norwegian. Bengali is distant from the above languages and uses a different script, whereas Norwegian is under-resourced but overlaps with English writing script to some extent.

\paragraph{LLM-as-a-judge}\label{sec:llm-as-a-judge}
To avoid expensive evaluation costs, we adopt LLM-as-a-judge \citep{llm-judge} to assign a score (1 to 3) to each instruction-response pair, and the final model score is the sum of its scores across all test instances. We use GPT-3.5 (\texttt{gpt-3.5-turbo-0613}) as the judge; it is queried with an instruction-response pair each time without model information or request history. 
We make modifications to \citet{llm-judge}'s prompt to ask the LLM to consider that an answer should be in the same language as the question, which is often the expectation with AI assistants.\footnote{There could be exceptions like text translation and code generation \citep{shaham2024multilingual}.} The exact wording is as \Cref{app:template} Figure~\ref{fig:gpt-eval-instruction}.

\paragraph{Language (in)consistency} Our manual inspection suggests that GPT-3.5 does not always obey the language requirement imposed. An example in \Cref{sec:appendix-language-inconsistency} \Cref{tab:example-language-inconsistency} shows a response in another language but scored highly. Hence, we run language identification and force-set a score to 0 if the response language is different from the query. We use the \texttt{fastText} framework \citep{joulin-etal-2017-bag} with \citet{burchell-etal-2023-open}'s checkpoint. The final response score can be framed as a product of GPT's quality score and a binary language identification outcome: $score = eval\_{score} \times lang\_id$. The aggregated test score thus ranges from 0 to 150.

\paragraph{Human-LLM agreement} We pick 600 outputs from 12 models to cover multilingual and monolingual systems and invite human evaluators to score each sample with an instruction similar to the LLM-as-a-judge prompt as in \Cref{app:human-template}. Four languages---English, Spanish, Bulgarian, and Chinese---are human-evaluated, and we obtain very high system-level Pearson correlation coefficients of 0.9225, 0.9683, 0.9205, and 0.8685, respectively between GPT-3.5 and human. Details are in \Cref{tab:human-evaluated-systems} in the appendix. This indicates the reliability of using LLM-as-a-judge to draw meaningful findings.

\begin{figure}[t]
\centering
\small
{\begin{minipage}[t]{0.25\textwidth}\centering
\input{plot_size_lora_bloom_on_en.tex}
\end{minipage}%
\begin{minipage}[t]{0.25\textwidth}\centering
\input{plot_size_lora_bloom_on_es.tex}
\end{minipage}\vspace{-1ex}}
{\begin{minipage}[t]{0.25\textwidth}\centering
\input{plot_size_lora_bloom_on_fr.tex}
\end{minipage}%
\begin{minipage}[t]{0.25\textwidth}\centering
\input{plot_size_lora_bloom_on_bg.tex}
\end{minipage}\vspace{-1ex}}
{\begin{minipage}[t]{0.25\textwidth}\centering
\input{plot_size_lora_bloom_on_ru.tex}
\end{minipage}%
\begin{minipage}[t]{0.25\textwidth}\centering
\input{plot_size_lora_bloom_on_zh.tex}
\end{minipage}}
\begin{minipage}[t]{0.5\textwidth}\centering
\input{plot_size_lora_bloom_legend.tex}
\end{minipage}
\caption{\textbf{LoRA} with \textbf{BLOOM} at different sizes. Caption: language; y-axis: score; x-axis: model size (B).}
\label{fig:lora-bloom-sizes}\vspace{-1ex}
\end{figure}
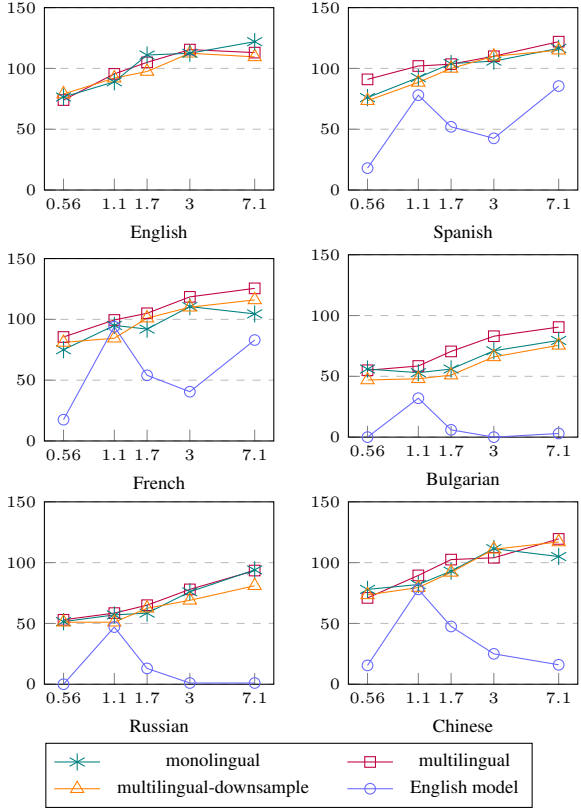

\section{Performance and Discussions}
\label{sec:results}

\subsection{Model sizes}
\label{sec:size}
\begin{figure}[t]\centering\small
{\begin{minipage}[t]{0.25\textwidth}\centering
\input{plot_size_fft_bloom_on_en.tex}
\end{minipage}%
\begin{minipage}[t]{0.25\textwidth}\centering
\input{plot_size_fft_bloom_on_es.tex}
\end{minipage}\vspace{-1ex}}
{\begin{minipage}[t]{0.25\textwidth}\centering
\input{plot_size_fft_bloom_on_fr.tex}
\end{minipage}%
\begin{minipage}[t]{0.25\textwidth}\centering
\input{plot_size_fft_bloom_on_bg.tex}
\end{minipage}\vspace{-1ex}}
{\begin{minipage}[t]{0.25\textwidth}\centering
\input{plot_size_fft_bloom_on_ru.tex}
\end{minipage}%
\begin{minipage}[t]{0.25\textwidth}\centering
\input{plot_size_fft_bloom_on_zh.tex}
\end{minipage}}
\caption{\textbf{FFT} with \textbf{BLOOM} at different sizes. Caption: language; y-axis: score; x-axis: model size (B). Same legend as \Cref{fig:lora-bloom-sizes}.
}
\label{fig:fft-bloom-sizes}\vspace{-1ex}
\end{figure}
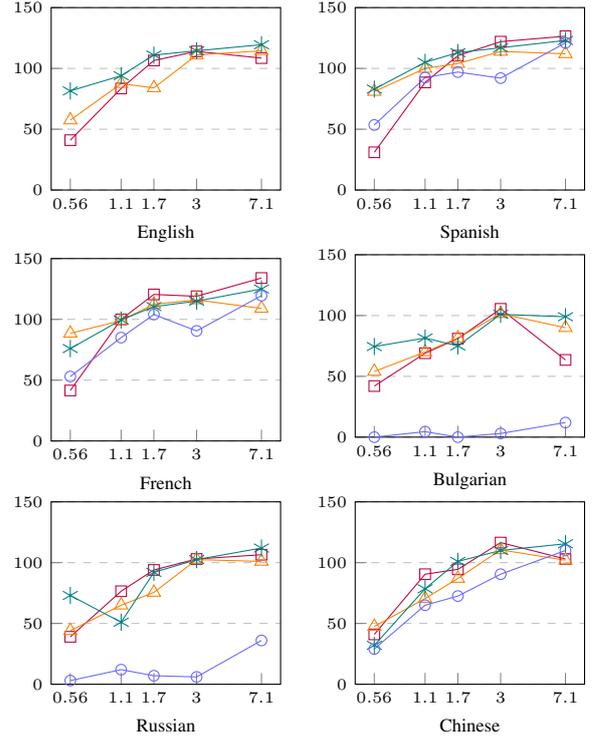

Results from LoRA fine-tuning of BLOOM at different sizes are shown in \Cref{fig:lora-bloom-sizes}. At smaller sizes, multilingual \begin{NoHyper}(\ref{tikz:patricks-purple-square-multilingual})\end{NoHyper} and monolingual \begin{NoHyper}(\ref{tikz:patricks-teal-asterisk-monolingual})\end{NoHyper} instruction tuning attain similar performance, and at larger sizes, multilingual models are generally better except for English. We observe similar trends for Pythia, placed in \Cref{app:model-size-pythia-lora} \Cref{fig:lora-pythia-sizes} due to space constraints. Moving on to full-parameter fine-tuning of BLOOM in \Cref{fig:fft-bloom-sizes}, we discover that at relatively small (<1.7B) or large sizes (7B), monolingual models are generally better than multilingual models for individual languages. 
\scalebox{0}{%
\begin{tikzpicture}
    \begin{axis}[hide axis]
        \addplot [mark=square,purple,semithick]
        (0,0);\label{tikz:patricks-purple-square-multilingual}
    \end{axis}%
    \begin{axis}[hide axis]
        \addplot [mark=asterisk,mark size=0.8ex,teal,semithick]
        (0,0);\label{tikz:patricks-teal-asterisk-monolingual}
    \end{axis}%
    \begin{axis}[hide axis]
        \addplot [mark=triangle,mark size=0.7ex,orange,semithick]
        (0,0);\label{tikz:patricks-orange-triangle-multilingual-downsample}
    \end{axis}%
    \begin{axis}[hide axis]
        \addplot [mark=o,blue!60,semithick]
        (0,0);\label{tikz:patricks-blue-o-english-model}
    \end{axis}%
\end{tikzpicture}%
}%
These observations suggest that multilingualism works well with LoRA, but separate monolingual tuning might be better with FFT. Overall, the LLMs' performance is correlated with sizes regardless of the tuning technique as anticipated.

\subsection{Budget-efficient tuning}
\label{sec:budgeting}
To aid our exploration of resource-constrained instruction tuning, in the aforementioned Figures~\ref{fig:lora-bloom-sizes}, ~\ref{fig:fft-bloom-sizes}, and ~\ref{fig:lora-pythia-sizes} (in \cref{app:model-size-pythia-lora}), we add the plots of two budget data conditions: using English-tuned models to respond to instructions in other languages \begin{NoHyper}(\ref{tikz:patricks-blue-o-english-model})\end{NoHyper}, as well as instruction tuning with downsampled multilingual data \begin{NoHyper}(\ref{tikz:patricks-orange-triangle-multilingual-downsample})\end{NoHyper}.

When using a single English model for all languages, its efficacy depends on the intended language/script's closeness to English: Spanish and French can maintain reasonable scores, but Bulgarian, Russian, and Chinese record very low performance. The only exception is BLOOM FFT in \Cref{fig:fft-bloom-sizes}, where the model is not too behind when operating in Chinese. Interestingly, BLOOM with LoRA sees a performance spike at 1.1B for non-English. At this specific size, it displayed multilingual transferability from pre-training and learned to follow multilingual instructions despite being fine-tuned merely in English.

In contrast, while consuming the same computational resources, downsampled multilingual tuning is significantly more robust across all test languages. These models sometimes achieve on-par performance with monolingual tuning in individual languages. This means that to support several languages with limited resources, the best practice is to train on small multilingual data even created with machine translation instead of full English data. Nonetheless, if the budget permits, training with the full multilingual data is still slightly better.

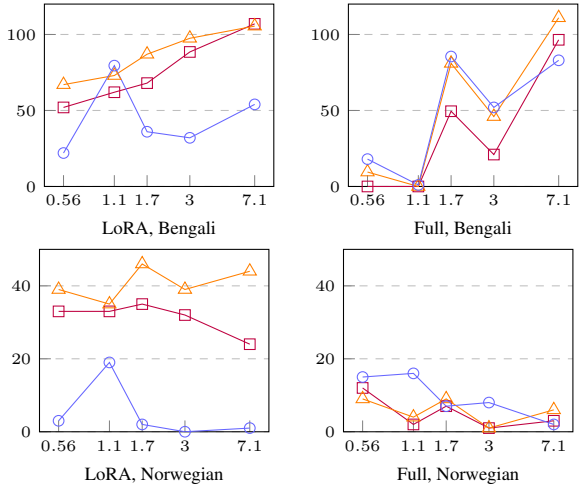
\begin{figure}[t]\centering\small
{\begin{minipage}[t]{0.25\textwidth}\centering
\input{plot_size_lora_bloom_on_bn.tex}
\end{minipage}%
\begin{minipage}[t]{0.25\textwidth}\centering
\input{plot_size_fft_bloom_on_bn.tex}
\end{minipage}}
{\begin{minipage}[t]{0.25\textwidth}\centering
\input{plot_size_lora_bloom_on_no.tex}
\end{minipage}%
\begin{minipage}[t]{0.25\textwidth}\centering
\input{plot_size_fft_bloom_on_no.tex}
\end{minipage}\vspace{-1ex}}
\caption{\textbf{LoRA} and \textbf{FFT} with \textbf{BLOOM} at different sizes and tested on \textbf{unseen} languages. Caption: training method and language; y-axis: score; x-axis: model size (B).}
\label{fig:lora-fft-bloom-sizes-unseen}\vspace{-1ex}
\end{figure}

\subsection{Unseen languages}
\label{sec:unseen}
Further in \Cref{fig:lora-fft-bloom-sizes-unseen}, we look at BLOOM models which underwent LoRA or FFT but were subsequently instructed in unseen languages at test time. English-tuned LLMs behave distinctly with LoRA and FFT. With the former, they are nowhere near multilingual tuned models, but with the latter, we see close or even better results. It might imply that FFT can even lift performance for languages not present in the instruction data. However, FFT results on Norwegian could be an outlier given its comparably low scores. Considering multilingual instruction tuning, we notice a pattern opposed to that on languages seen during training---learning on the downsampled data is superior to ingesting the full mixed data. We conclude that it is important to not overfit to instruction languages if unseen languages are expected in downstream tasks.

\subsection{Language robustness}
\label{sec:language-consistency}
\begin{figure}[t]\centering\small
{\begin{minipage}[t]{0.25\textwidth}\centering
\vfill
\input{plot_robustness_lora_bloom.tex}
\end{minipage}%
\begin{minipage}[t]{0.25\textwidth}\centering
\vfill
\input{plot_robustness_lora_pythia.tex}
\end{minipage}}
{\begin{minipage}[t]{0.25\textwidth}\centering
\vfill\vspace{-1ex}
\input{plot_robustness_fft_bloom.tex}
\end{minipage}%
\begin{minipage}[t]{0.25\textwidth}\centering
\vfill
\input{plot_robustness_lora_fft_bloom_pythia_legend.tex}
\end{minipage}}
\caption{Evaluation \textbf{score change} before and after language identification, \textbf{averaged} over six seen test languages, at different LLM sizes. Caption: training method and base model; y-axis: score difference (log scale); x-axis: model size (B).}
\label{fig:lora-bloom-pythia-sizes-robustness}\vspace{-1ex}
\end{figure}
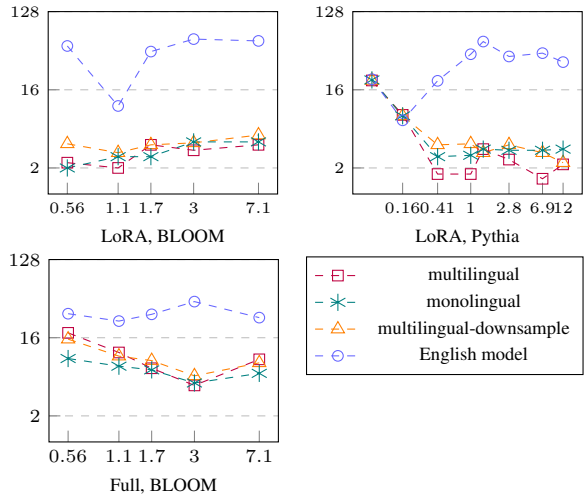

We review each model and data recipe's scores before and after adding language identification, to isolate an LLM's language robustness from its ``inherent quality'' (regardless of the response language). 
We compute the \textit{differences} in GPT evaluation scores before and after applying language identification. A (big) difference suggests that a model produces reasonable answers in an undesired language. In \Cref{fig:lora-bloom-pythia-sizes-robustness}, we report the \textit{average} of the score differences across all six test languages seen during tuning. English-only models are the least robust---their score differences are way above other techniques. With LoRA, full multilingual tuning records the smallest performance drop; with FFT, monolingual tuning is preferred. The insights from language robustness are corroborated by our early findings in \Cref{sec:size}: superior results are obtained when using multilingual tuning with LoRA and monolingual tuning with full-parameter tuning. Nonetheless, monolingual and multilingual tuning are not too far apart; specifically for BLOOM with LoRA, language robustness does not improve as the model gets larger.

\subsection{Model families}
\begin{figure*}[t]\centering\small
\begin{minipage}[t]{0.33\textwidth}\centering
\input{plot_family_lora_7b_on_en.tex}
\end{minipage}%
\begin{minipage}[t]{0.33\textwidth}\centering
\input{plot_family_lora_7b_on_es.tex}
\end{minipage}%
\begin{minipage}[t]{0.33\textwidth}\centering
\input{plot_family_lora_7b_on_fr.tex}
\end{minipage}
\begin{minipage}[t]{0.33\textwidth}\centering
\input{plot_family_lora_7b_on_bg.tex}
\end{minipage}%
\begin{minipage}[t]{0.33\textwidth}\centering
\input{plot_family_lora_7b_on_ru.tex}
\end{minipage}%
\begin{minipage}[t]{0.33\textwidth}\centering
\input{plot_family_lora_7b_on_zh.tex}
\end{minipage}
\begin{minipage}[t]{1\textwidth}\centering
\input{plot_family_lora_7b_legend.tex}
\end{minipage}
\caption{LoRA fine-tuning on different 7B LLMs. Caption: language generated; y-axis: score; x-axis: model family.}\label{fig:lora-7b-family}
\end{figure*}
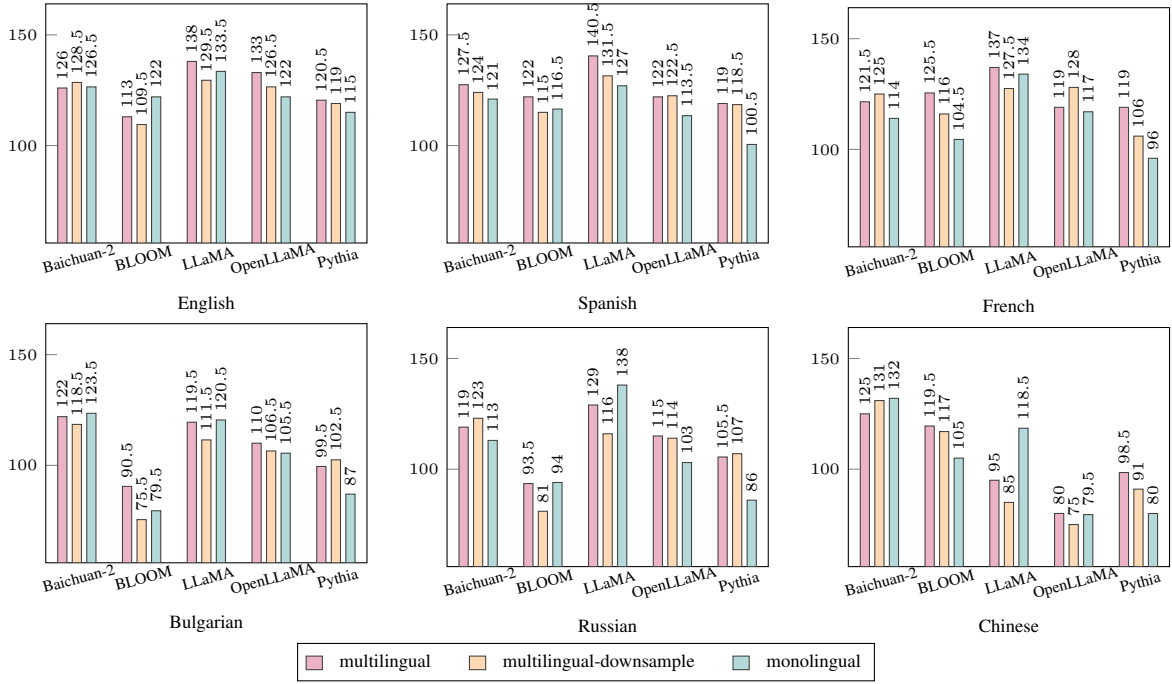

Finally, we experiment with base LLMs from different families of around 7 billion parameters. In \Cref{fig:lora-7b-family}, we plot the evaluation scores for multilingual, downsampled multilingual, and monolingual LoRA tuning for six languages. Generally, LLaMA and OpenLLaMA have better performance than BLOOM and Pythia potentially because they have pre-training data that is an order of magnitude larger. Also Bulgarian, Russian, and Chinese see lower scores than English, again presumably due to the language distribution in the pre-training data.

Delving into the comparison between monolingual and multilingual instruction tuning, we find that out of 30 cases across six languages and five LLMs, monolingual tuning is ahead in just two cases: LLaMA tested in Russian and Chinese. The cost-efficient downsampled multilingual tuning leads in four cases: two in French and two in Russian. In other situations, multilingual training is on par if not better. The outcome of tuning several similar-sized LLMs confirms that multilingual tuning is favourable using LoRA.

\section{Related Work}
\label{sec:related-work}

Many large language models appeared recently: the closed-source GPT model family \citep{radford2019language,brown2020language,ouyang2022training}; open-source English-centric models like LLaMA \citep{touvron2023llama}, OpenLLaMA \citep{openlm2023openllama}, and Pythia \citep{biderman2023pythia}; open-source multilingual models like mT5 \citep{xue-etal-2021-mt5} and BLOOM \citep{scao2022bloom}. These models have exhibited different degrees of language versatility.

LLM pre-training data is usually skewed towards English. One way to improve an LLM's coverage of non-English languages is through continued pre-training \citep[][inter alia]{chinese_alpaca}. Another rich body of literature looks into multilingualism in instruction tuning, which is used to adjust base models to respond to input \cite{mishra-etal-2022-cross,sanh2021multitask,wei2022finetuned,longpre2023flan}. It trains an LLM by providing downstream tasks' input and output in a specific format. Early research created a multilingual instruction dataset using machine translation and showed that multilingual tuning gained higher performance than English-only fine-tuning \citep{muennighoff-etal-2023-crosslingual-xp3}. They also found that low-cost translated instructions are superior to human-written non-English prompts on multiple language understanding tasks.

Lately, multiple contemporaneous papers delving into multilingual instruction tuning have been made public on arXiv---some appeared before our work and some after. This reflects the importance and interest in widening LLMs' language support. \citet{li2023bactrian} created an instruction dataset with instructions translated from English but responses generated by an LLM. When tuned with LoRA, their monolingual models outperform multilingual ones on language understanding tasks. \citet{wei2023polylm} created a multilingual counterpart of Alpaca using self-instruct. It has also been showcased that translation instructions improve cross-lingual capabilities \citep{li2023eliciting,BayLing,ranaldi2023empowering} and research explored more cross-lingual task data and multilingual tuning \citep{zhu2023extrapolating}. Moreover, researchers have unveiled that fine-tuning on a modest number of languages---approximately three---seems to effectively instigate cross-lingual transfer in downstream tasks \citep{kew2023turning,shaham2024multilingual}.

\FloatBarrier
\section{Conclusion}
This paper presents a study of instruction tuning of large language models in different language contexts. Our study in a resource-controlled setting suggests that multilingual tuning offers more benefits compared to monolingual tuning. We find that multilingual tuning on a downsampled dataset achieves better robustness on unseen languages.

\section*{Limitations}
The LLMs we studied have primarily 7B and at most 13B parameters and the multilingual training only spanned nine languages. Scaling to larger models and more languages would be interesting. The best checkpoint for our instruction fine-tuning is selected based on validation cross-entropy, but there is no guarantee that this leads to the best performance on the downstream task.

To manage the budget for human translation and evaluation, we consider eight languages (six seen and two unseen languages during instruction tuning) to translate and sample 50 instances for evaluation. The training data for non-English languages are obtained via machine translation, which introduces errors, affects response fluency, and might alter the nature of some tasks such as grammatical error correction and code generation.

\section*{Ethics Statement}
The dataset we translated and generated does not contain private or sensitive information. Similar to other research on large language models, there is no definitive way for us to prevent the instruction-tuned models from generating inappropriate content. However, we see minimal such risks associated with our project, as neither our models nor generated contents are intended for public consumption. Human evaluators did not report inappropriate content generated by the models.

\section*{Acknowledgements}
This paper stemmed from a hackathon project organized by the High Performance Language Technologies (HPLT) consortium.\footnote{\url{https://hplt-project.org}} We are grateful to Alicia N\'{u}\~{n}ez Alcover, David Samuel, Joona Kyt\"{o}niemi, J\"{o}rg Tiedemann, Lucas Charpentier, Sampo Pyysalo, Petter M\ae{}hlum, and Zhicheng Guo for project discussions, test data translation, and evaluation setup.

The work has received funding from the European Union's Horizon Europe research and innovation programme under grant agreement No 101070350, from UK Research and Innovation (UKRI) under the UK government’s Horizon Europe funding guarantee [grant number 10052546], as well as from the European Research Council (ERC) under the EU's Horizon 2020 research and innovation program (agreement \textnumero{}~771113).

Computation in this work was performed on LUMI, Karolina, and Baskerville. 
We acknowledge CSC-IT Center for Science, Finland for awarding this project access to the LUMI supercomputer, owned by the EuroHPC Joint Undertaking, hosted by CSC (Finland) and the LUMI consortium through Finnish extreme scale call (project LumiNMT). 
Karolina was supported by the Ministry of Education, Youth and Sports of the Czech Republic through the e-INFRA CZ (ID:90254). 
The Baskerville Tier 2 HPC was funded by the EPSRC and UKRI through the World Class Labs scheme (EP/T022221/1) and the Digital Research Infrastructure programme (EP/W032244/1) and is operated by Advanced Research Computing at the University of Birmingham.

\bibliography{bibliography/custom,bibliography/anthology_1,bibliography/anthology_2}

\appendix
\section{Experimental Setup Details}

\subsection{Hyperparameters}
\label{app:exp-setup}

\Cref{tab:hyperparameters} shows the hyperparameter configurations of LoRA and full-parameter fine-tuning. LoRA is a parameter-efficient training method where, for a big matrix, only low-rank matrices are trained and patched to it. In our case, we apply it to the attention matrices (key, query, value) and use rank 8, dropout 0.05, and scaling factor 16 throughout. We use a batch size of 128, set a fixed training budget of 5 epochs with a learning rate of 3e\textsuperscript{-4}, and select the best checkpoint based on validation cross-entropy. For full-parameter fine-tuning, we follow the configurations of Alpaca by training for 3 epochs with a learning rate of 2e\textsuperscript{-5}, a warm-up ratio of 0.03, and a batch size of 256.

Since we use a range of models of different sizes, we estimate computation time based on 7-billion parameter models which are the second largest we fine-tuned. LoRA tuning takes 15-20 hours on 4 GeForce RTX 3090 GPUs, using CPU memory offloading and distributed training. Full-parameter fine-tuning is performed on 4 AMD MI250x GPUs (treated as 8 GPUs with 64G memory each at runtime) with model parallelism, and it requires around 24 hours to finish. Given the high computational cost of model fine-tuning, we conducted all fine-tuning experiments once. We use a range of different GPUs, but through gradient accumulation, we maintain the same global batch size for each tuning technique: 128 for LoRA and 256 for full-parameter fine-tuning.

\input{table_hyperparameters.tex}

\subsection{Description of LLMs}
\label{app:exp-base-model}
Due to the space constraint, we place a detailed description of LLMs used in our research here. All the models used in this study are publicly available and free to use for academic purposes.

\textbf{Baichuan-2}~\citep{yang2023baichuan} is a multilingual LLM trained on 2.6 trillion tokens. While the data composition is not transparent in its technical report, the LLM weights are open-source and it performs strongly on tasks in English and Chinese. We use its 7B checkpoint.
    
\textbf{BLOOM}~\citep{scao2022bloom} is trained on the ROOTS dataset \citep{laurenccon2022roots} containing 350 billion tokens in 46 natural languages spanning 9 language families and 12 programming languages. The LLM has English, Chinese, French, and Spanish as the major components. We use the checkpoints from 560M to 7.1B for experiments.
    
\textbf{LLaMA}~\citep{touvron2023llama} has been trained on data mainly in English with some in European languages in Latin and Cyrillic scripts. It could also support other languages with byte-BPE tokenization. We use its 7B model which has seen 1 trillion tokens.
    
\textbf{OpenLLaMA}~\cite{openlm2023openllama} is an open-source reproduction of LLaMA, trained on the RedPajama dataset \citep{together2023redpajama}, which is close to LLaMA's data composition. Similarly, we use the 7B version.
    
\textbf{Pythia}~\citep{biderman2023pythia} is trained on the Pile \citep{gao2020pile} which has nearly 300 billion tokens and is intended to be English-dominant. We experiment with the full range from 70M to 12B.

\section{Evaluation Details}

\subsection{Prompt template for LLM-as-a-judge}
\label{app:template}

Our LLM-as-a-judge prompt template is as \Cref{fig:gpt-eval-instruction}, similar to \citet{llm-judge}'s with an additional requirement on the response language.

\begin{figure}[ht]
    \centering\small
    \noindent\framebox{%
    \parbox{0.47\textwidth}{
    \texttt{Please act as an impartial judge and evaluate the quality of the response provided by an AI assistant to the user instruction displayed below. Your evaluation should consider factors such as helpfulness, relevance, accuracy, depth, creativity, and level of detail. It is also required that the response is in the same language as the instruction. Begin your evaluation with a short explanation. Be as objective as possible. After providing your explanation, please rate the response on a scale of 1 to 3 by strictly following this format:``[[rating]]'', for example: ``Rating: [[2]]''.}\\
    
    \texttt{[User instruction]}
    
    \texttt{\$\{instruction\}}
    \\
    
    \texttt{[AI assistant's response]}
    
    \texttt{\$\{response\}}
    }%
    }
\caption{Prompt template for requesting a response evaluation from GPT-3.5, where \texttt{\$\{instruction\}} and \texttt{\$\{response\}} are replaced with questions and model responses.}
\label{fig:gpt-eval-instruction}
\end{figure}

\subsection{An example of language inconsistency}
\label{sec:appendix-language-inconsistency}

\Cref{tab:example-language-inconsistency} lists an example where the response is ``correct'' in content but does not match the language of the query. In this case, the judge, GPT-3.5 does not penalize this language inconsistency despite being asked to do so in the evaluation prompt. This makes extra language identification essential.

\begin{table}[ht]
\centering\small
\begin{tabularx}{0.48\textwidth}{X}
\toprule
   \textbf{Instruction:} \begin{CJK*}{UTF8}{gbsn}你好！ 你能帮我计划婚礼吗？\end{CJK*} \\
   \textit{(English: Hello! Can you help me plan my wedding?)} \\
\cdashlinelr{1-1}
   \textbf{Response:} Sure, I would be happy to help you plan your wedding. Can you tell me a little bit more about your wedding plans and what you are looking for in a wedding planner? \\
\midrule
   \textbf{GPT-3.5 explanation:} The AI assistant's response shows a good understanding of the user's request and offers assistance. It asks the user for more information about their wedding plans, which shows that it is trying to gather relevant details to provide appropriate help. The response is polite and friendly, which contributes to a positive user experience. \\
\cdashlinelr{1-1}
   \textbf{GPT-3.5 score:} 3 \\
\bottomrule
\end{tabularx}
\caption{An example of instruction-response language inconsistency, where the question is in Chinese but the response is in English, which GPT-3.5 fails to account for in its evaluation.} 
\label{tab:example-language-inconsistency}
\end{table}

\begin{table*}[t]
\centering\small
\begin{tabular}{ccccccccccc}
\toprule
& \multirow{2}{*}{LLM} & \multirow{2}{*}{\makecell{Size\\(B)}} & \multicolumn{2}{c}{English}  & \multicolumn{2}{c}{Spanish}  & \multicolumn{2}{c}{Bulgarian}  & \multicolumn{2}{c}{Chinese}     \\
\cmidrule(lr){4-5}\cmidrule(lr){6-7}\cmidrule(lr){8-9}\cmidrule(lr){10-11}
 & &  & GPT-3.5 & human & GPT-3.5 & human & GPT-3.5 & human & GPT-3.5 & human\\
\midrule
\multirow{6}{*}{\makecell{Multi-\\lingual}} & BLOOM  & 1.1 & \phantom{0}95.5  & \phantom{0}93.0 & 102.0 & \phantom{0}98.0  & \phantom{0}58.5  & \phantom{0}54.5 & \phantom{0}89.5  & \phantom{0}97.5 \\
 & BLOOM  & 3  & 115.5 & 105.0  & 110.0 & 103.5 & \phantom{0}83.0  & \phantom{0}59.0  & 104.0 & 102.0   \\
 & BLOOM  & 7.1 & 113.0 & 119.5 & 122.0 & 116.5 & \phantom{0}90.5  & \phantom{0}67.0 & 119.5 & 117.5\\
 & LLaMA  & 7  & 138.0 & 131.5 & 140.5 & 123.0 & 119.5 & 112.0 & \phantom{0}95.0  & \phantom{0}89.0    \\
 & OpenLLaMA & 7  & 133.0 & 130.0  & 122.0 & 112.5 & 110.0 & \phantom{0}89.0  & \phantom{0}80.0  & \phantom{0}67.5 \\
 & Pythia & 6.9 & 120.5 & 117.0 & 119.0 & 107.5 & \phantom{0}99.5  & \phantom{0}75.0  & \phantom{0}98.5  & \phantom{0}87.5 \\
\cdashlinelr{1-11}
\multirow{6}{*}{\makecell{Mono-\\lingual}} & BLOOM  & 1.1 & \phantom{0}89.0  & \phantom{0}81.0  & \phantom{0}92.5  & \phantom{0}86.0  & \phantom{0}53.0  & \phantom{0}49.0  & \phantom{0}82.0  & \phantom{0}75.5 \\
 & BLOOM  & 3  & 112.5 & 103.5 & 106.0 & \phantom{0}99.5 & \phantom{0}71.0  & \phantom{0}64.0  & 111.5 & \phantom{0}96.0    \\
 & BLOOM  & 7.1 & 122.0 & 111.5 & 116.5 & 111.5 & \phantom{0}79.5  & \phantom{0}73.5 & 105.0 & 106.0   \\
 & LLaMA  & 7  & 133.5 & 121.0 & 127.0 & 115.0 & 120.5 & 117.5 & 118.5 & \phantom{0}96.5 \\
 & OpenLLaMA & 7  & 122.0 & 124.0 & 113.5 & 108.0  & 105.5 & \phantom{0}87.0  & \phantom{0}79.5  & \phantom{0}66.5 \\
 & Pythia & 6.9 & 115.0 & 116.0 & 100.5 & \phantom{0}97.5 & \phantom{0}87.0  & \phantom{0}72.5 & \phantom{0}80.0  & \phantom{0}72.0    \\
\cmidrule(r){1-3}\cmidrule(lr){4-5}\cmidrule(lr){6-7}\cmidrule(lr){8-9}\cmidrule(lr){10-11}
\multicolumn{3}{c}{Pearson correlation coefficient} & \multicolumn{2}{c}{\textbf{0.9225}} & \multicolumn{2}{c}{\textbf{0.9683}} & \multicolumn{2}{c}{\textbf{0.9205}} & \multicolumn{2}{c}{\textbf{0.8685}} \\
\bottomrule
\end{tabular}
\caption{Human evaluation scores and their system-level correlation with GPT-3.5 scores. Models are fine-tuned with LoRA.}
\label{tab:human-evaluated-systems}
\end{table*}

\subsection{Human evaluation and human-LLM agreement}
\label{app:human-template}

We invited human evaluators who are fluent or native in the language of the instructions and responses to score in total outputs from 12 models fine-tuned with LoRA. We attach the instruction given to human evaluators in \Cref{fig:human-eval-instruction}. The systems' responses for the same instruction are shuffled but grouped together to provide a context of the overall quality. The human evaluators are asked to assign each response a score. We list the model details, as well as their aggregated GPT and human evaluation scores in \Cref{tab:human-evaluated-systems}.

\begin{figure}[ht]
    \centering\small
    \noindent\framebox{%
    \parbox{0.47\textwidth}{
    \texttt{Please evaluate the quality of the responses provided by AI assistants to the questions in your respective tab. Most questions are open-ended, meaning there is no strictly correct or best answer. Please make a judgment based on your perspective of quality. You could consider factors such as helpfulness, relevance, accuracy, depth, creativity, and level of detail. It is also required that the response is in the same language as the question unless otherwise specified by the instruction itself. Please rate the response on a scale of 0 to 3. If you feel indecisive, you can use an increment of 0.5. You can give a score of 0 for ``incorrect language, not readable, content cannot be understood''; give a score of 1 for ``a relatively bad response''; give a score of 2 for ``a medium response''; give a score of 3 for ``a relatively good response''.}
    }%
    }
\caption{Instructions for human evaluators.}
\label{fig:human-eval-instruction}
\end{figure}

\section{Result Details}

\subsection{Experiments on Pythia with LoRA}
\label{app:model-size-pythia-lora}
\begin{figure}[ht]
\centering\small
{\begin{minipage}[t]{0.25\textwidth}\centering
\input{plot_size_lora_pythia_on_en.tex}
\end{minipage}%
\begin{minipage}[t]{0.25\textwidth}\centering
\input{plot_size_lora_pythia_on_es.tex}
\end{minipage}\vspace{-1ex}}
{\begin{minipage}[t]{0.25\textwidth}\centering
\input{plot_size_lora_pythia_on_fr.tex}
\end{minipage}%
\begin{minipage}[t]{0.25\textwidth}\centering
\input{plot_size_lora_pythia_on_bg.tex}
\end{minipage}\vspace{-1ex}}
{\begin{minipage}[t]{0.25\textwidth}\centering
\input{plot_size_lora_pythia_on_ru.tex}
\end{minipage}%
\begin{minipage}[t]{0.25\textwidth}\centering
\input{plot_size_lora_pythia_on_zh.tex}
\end{minipage}}
\begin{minipage}[t]{0.5\textwidth}\centering
\input{plot_size_lora_pythia_legend.tex}
\end{minipage}
\caption{\textbf{LoRA} fine-tuning on \textbf{Pythia}. 
Caption: language generated; y-axis: score; x-axis: model size (B) on a logarithmic scale.}
\label{fig:lora-pythia-sizes}
\end{figure}
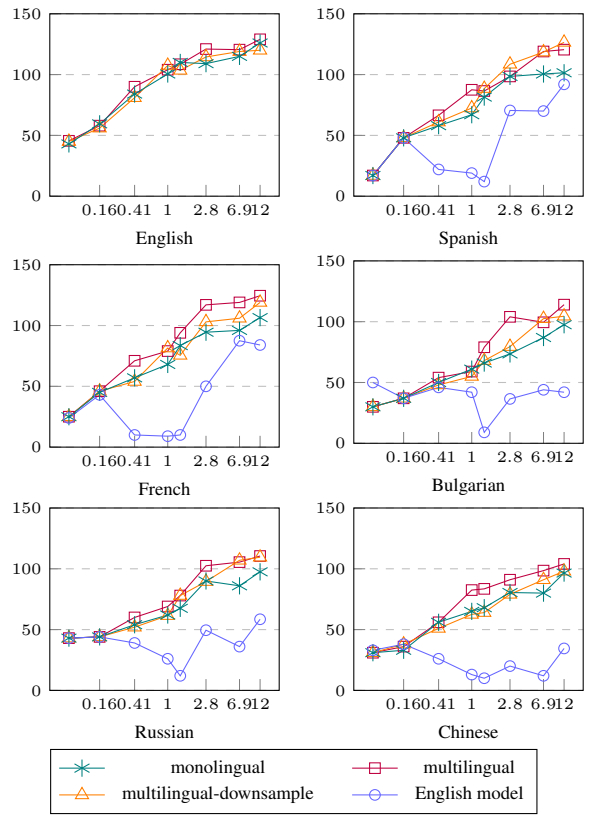

Apart from LoRA fine-tuning on BLOOM models, we conduct the same investigation on Pythia models at different sizes. We observe that multilingual tuning does not lose to monolingual tuning in any language, similar to what we find about BLOOM in \Cref{sec:size}. The plots for the six languages are included as \Cref{fig:lora-pythia-sizes}.

\end{document}

%% file: plot_size_lora_bloom_on_en.tex
\begin{tikzpicture}
\begin{axis}[
 xlabel=English,
 width=1.15\linewidth,height=\linewidth,
 ymax=150, ymin=0,
 xtick={0.56, 1.1, 1.7, 3, 7.1},
 xmode=log, log basis x={2},
 xticklabel=\pgfmathparse{2^\tick}\pgfmathprintnumber{\pgfmathresult},
 yticklabel style={
        /pgf/number format/fixed,
        /pgf/number format/fixed zerofill,
        /pgf/number format/precision=0
 },
 tick label style={font=\tiny},
 ymajorgrids=true,
 grid style=dashed,
 xlabel style={font=\scriptsize}
 ]
  \addplot[mark=square,
           purple,
           ] coordinates {
            (0.56, 74.0)
            (1.1, 95.5)
            (1.7, 105.0)
            (3, 115.5)
            (7.1, 113.0)};
    \addplot[mark=triangle, mark size=0.7ex,
           orange,
           ] coordinates {
            (0.56, 79.0)
            (1.1, 92.0)
            (1.7, 97.5)
            (3, 112.5)
            (7.1, 109.5)};
    \addplot[mark=asterisk, mark size=0.8ex,
           teal,
           ] coordinates {
            (0.56, 76.5)
            (1.1, 89.0)
            (1.7, 111.0)
            (3, 112.5)
            (7.1, 122.0)};
\end{axis}
\end{tikzpicture}

%% file: plot_size_lora_bloom_on_es.tex
\begin{tikzpicture}
\begin{axis}[
 xlabel=Spanish,
 width=1.15\linewidth,height=\linewidth,
 ymax=150, ymin=0,
 xtick={0.56, 1.1, 1.7, 3, 7.1},
 xmode=log, log basis x={2},
 xticklabel=\pgfmathparse{2^\tick}\pgfmathprintnumber{\pgfmathresult},
 yticklabel style={
        /pgf/number format/fixed,
        /pgf/number format/fixed zerofill,
        /pgf/number format/precision=0
 },
 tick label style={font=\tiny},
 ymajorgrids=true,
 grid style=dashed,
 xlabel style={font=\scriptsize}
 ]
  \addplot[mark=square,
           purple,
           ] coordinates {
            (0.56, 91.0)
            (1.1, 102.0)
            (1.7, 103.5)
            (3, 110.0)
            (7.1, 122.0)};
    \addplot[mark=asterisk, mark size=0.8ex,
           teal,
           ] coordinates {
            (0.56, 76.0)
            (1.1, 92.5)
            (1.7, 104.0)
            (3, 106.0)
            (7.1, 116.5)};
    \addplot[mark=triangle, mark size=0.7ex,
           orange,
           ] coordinates {
            (0.56, 73.5)
            (1.1, 88.5)
            (1.7, 100.0)
            (3, 110.0)
            (7.1, 115.0)};
    \addplot[mark=o,
             blue!60,
           ] coordinates {
            (0.56, 18)
            (1.1, 78)
            (1.7, 52.0)
            (3, 42.5)
            (7.1, 85.5)};
\end{axis}
\end{tikzpicture}

%% file: plot_size_lora_bloom_on_fr.tex
\begin{tikzpicture}
\begin{axis}[
 xlabel=French,
 width=1.15\linewidth,height=\linewidth,
 ymax=150, ymin=0,
 xtick={0.56, 1.1, 1.7, 3, 7.1},
 xmode=log, log basis x={2},
 xticklabel=\pgfmathparse{2^\tick}\pgfmathprintnumber{\pgfmathresult},
 yticklabel style={
        /pgf/number format/fixed,
        /pgf/number format/fixed zerofill,
        /pgf/number format/precision=0
 },
 tick label style={font=\tiny},
 ymajorgrids=true,
 grid style=dashed,
 xlabel style={font=\scriptsize}
 ]
  \addplot[mark=square,
           purple,
           ] coordinates {
            (0.56, 85.5)
            (1.1, 99.5)
            (1.7, 105.0)
            (3, 118.5)
            (7.1, 125.5)};
    \addplot[mark=asterisk, mark size=0.8ex,
           teal,
           ] coordinates {
            (0.56, 75.0)
            (1.1, 95.0)
            (1.7, 92.0)
            (3, 110.5)
            (7.1, 104.5)};
    \addplot[mark=triangle, mark size=0.7ex,
           orange,
           ] coordinates {
            (0.56, 81.0)
            (1.1, 84.5)
            (1.7, 101.0)
            (3, 110.0)
            (7.1, 116.0)};
    \addplot[mark=o,
             blue!60,
           ] coordinates {
            (0.56, 17.5)
            (1.1, 94)
            (1.7, 54)
            (3, 40.5)
            (7.1, 83)};
\end{axis}
\end{tikzpicture}

%% file: plot_size_lora_bloom_on_bg.tex
\begin{tikzpicture}
\begin{axis}[
 xlabel=Bulgarian,
 width=1.15\linewidth,height=\linewidth,
 ymax=150, ymin=0,
 xtick={0.56, 1.1, 1.7, 3, 7.1},
 xmode=log, log basis x={2},
 xticklabel=\pgfmathparse{2^\tick}\pgfmathprintnumber{\pgfmathresult},
 yticklabel style={
        /pgf/number format/fixed,
        /pgf/number format/fixed zerofill,
        /pgf/number format/precision=0
 },
 tick label style={font=\tiny},
 ymajorgrids=true,
 grid style=dashed,
 xlabel style={font=\scriptsize}
 ]
  \addplot[mark=square,
           purple,
           ] coordinates {
            (0.56, 55.0)
            (1.1, 58.5)
            (1.7, 70.5)
            (3, 83.0)
            (7.1, 90.5)};
    \addplot[mark=asterisk, mark size=0.8ex,
           teal,
           ] coordinates {
            (0.56, 56.0)
            (1.1, 53.0)
            (1.7, 56.0)
            (3, 71.0)
            (7.1, 79.5)};
    \addplot[mark=triangle, mark size=0.7ex,
           orange,
           ] coordinates {
            (0.56, 47.0)
            (1.1, 48.0)
            (1.7, 51.0)
            (3, 66.0)
            (7.1, 75.5)};
    \addplot[mark=o,
             blue!60,
           ] coordinates {
            (0.56, 0)
            (1.1, 32)
            (1.7, 6)
            (3, 0)
            (7.1, 3)};
\end{axis}
\end{tikzpicture}

%% file: plot_size_lora_bloom_on_ru.tex
\begin{tikzpicture}
\begin{axis}[
 xlabel=Russian,
 width=1.15\linewidth,height=\linewidth,
 ymax=150, ymin=0,
 xtick={0.56, 1.1, 1.7, 3, 7.1},
 xmode=log, log basis x={2},
 xticklabel=\pgfmathparse{2^\tick}\pgfmathprintnumber{\pgfmathresult},
 yticklabel style={
        /pgf/number format/fixed,
        /pgf/number format/fixed zerofill,
        /pgf/number format/precision=0
 },
 tick label style={font=\tiny},
 ymajorgrids=true,
 grid style=dashed,
 xlabel style={font=\scriptsize}
 ]
  \addplot[mark=square,
           purple,
           ] coordinates {
            (0.56, 53.0)
            (1.1, 58.5)
            (1.7, 65.0)
            (3, 78.0)
            (7.1, 93.5)};
    \addplot[mark=asterisk, mark size=0.8ex,
           teal,
           ] coordinates {
            (0.56, 51.5)
            (1.1, 57.0)
            (1.7, 58.5)
            (3, 76.0)
            (7.1, 94.0)};
    \addplot[mark=triangle, mark size=0.7ex,
           orange,
           ] coordinates {
            (0.56, 51.0)
            (1.1, 51.0)
            (1.7, 62.5)
            (3, 69.0)
            (7.1, 81.0)};
    \addplot[mark=o,
             blue!60,
           ] coordinates {
            (0.56, 0)
            (1.1, 47)
            (1.7, 13)
            (3, 1)
            (7.1, 1)};
\end{axis}
\end{tikzpicture}

%% file: plot_size_lora_bloom_on_zh.tex
\begin{tikzpicture}
\begin{axis}[
 xlabel=Chinese,
 width=1.15\linewidth,height=\linewidth,
 ymax=150, ymin=0,
 xtick={0.56, 1.1, 1.7, 3, 7.1},
 xmode=log, log basis x={2},
 xticklabel=\pgfmathparse{2^\tick}\pgfmathprintnumber{\pgfmathresult},
 yticklabel style={
        /pgf/number format/fixed,
        /pgf/number format/fixed zerofill,
        /pgf/number format/precision=0
 },
 tick label style={font=\tiny},
 ymajorgrids=true,
 grid style=dashed,
 xlabel style={font=\scriptsize}
 ]
  \addplot[mark=square,
           purple,
           ] coordinates {
            (0.56, 71.0)
            (1.1, 89.5)
            (1.7, 102.5)
            (3, 104.0)
            (7.1, 119.5)};
    \addplot[mark=asterisk, mark size=0.8ex,
           teal,
           ] coordinates {
            (0.56, 78.0)
            (1.1, 82.0)
            (1.7, 93.0)
            (3, 111.5)
            (7.1, 105.0)};
    \addplot[mark=triangle, mark size=0.7ex,
           orange,
           ] coordinates {
            (0.56, 73.5)
            (1.1, 79.5)
            (1.7, 92.0)
            (3, 111.0)
            (7.1, 117.0)};
    \addplot[mark=o,
             blue!60,
           ] coordinates {
            (0.56, 15.5)
            (1.1, 78.0)
            (1.7, 47.5)
            (3, 25.0)
            (7.1, 16.0)};
\end{axis}
\end{tikzpicture}

%% file: plot_size_lora_bloom_legend.tex
\begin{tikzpicture}
\begin{customlegend}[
    legend columns=2,
    legend style={
            column sep=3ex,
            font=\scriptsize,
    },
    legend entries={
            \hspace{-2.5ex}monolingual,
            \hspace{-2.5ex}multilingual,
            \hspace{-2.5ex}multilingual-downsample,
            \hspace{-2.5ex}English model,
    }
]
\addlegendimage{
    mark=asterisk,
    mark size=0.8ex,
    teal,
}
\addlegendimage{
    mark=square,
    purple,
}
\addlegendimage{
    mark=triangle,
    mark size=0.7ex,
    orange,
}
\addlegendimage{
    mark=o,
    blue!60,
}
\end{customlegend}
\end{tikzpicture}

%% file: plot_size_fft_bloom_on_en.tex
\begin{tikzpicture}
\begin{axis}[
 xlabel=English,
 width=1.15\linewidth,height=\linewidth,
 ymax=150, ymin=0,
 xtick={0.56, 1.1, 1.7, 3, 7.1},
 xmode=log, log basis x={2},
 xticklabel=\pgfmathparse{2^\tick}\pgfmathprintnumber{\pgfmathresult},
 yticklabel style={
        /pgf/number format/fixed,
        /pgf/number format/fixed zerofill,
        /pgf/number format/precision=0
 },
 tick label style={font=\tiny},
 ymajorgrids=true,
 grid style=dashed,
 xlabel style={font=\scriptsize}
 ]
  \addplot[mark=square,
           purple,
           ] coordinates {
            (0.56, 41.0)
            (1.1, 83.5)
            (1.7, 106.5)
            (3, 114.0)
            (7.1, 108.5)};
    \addplot[mark=triangle, mark size=0.7ex,
           orange,
           ] coordinates {
            (0.56, 57.5)
            (1.1, 87.5)
            (1.7, 84.0)
            (3, 111.0)
            (7.1, 114.5)};
    \addplot[mark=asterisk, mark size=0.8ex, teal,
           ] coordinates {
            (0.56, 81.5)
            (1.1, 94.0)
            (1.7, 111.0)
            (3, 114.5)
            (7.1, 119.5)};
\end{axis}
\end{tikzpicture}

%% file: plot_size_fft_bloom_on_es.tex
\begin{tikzpicture}
\begin{axis}[
 xlabel=Spanish,
 width=1.15\linewidth,height=\linewidth,
 ymax=150, ymin=0,
 xtick={0.56, 1.1, 1.7, 3, 7.1},
 xmode=log, log basis x={2},
 xticklabel=\pgfmathparse{2^\tick}\pgfmathprintnumber{\pgfmathresult},
 yticklabel style={
        /pgf/number format/fixed,
        /pgf/number format/fixed zerofill,
        /pgf/number format/precision=0
 },
 tick label style={font=\tiny},
 ymajorgrids=true,
 grid style=dashed,
 xlabel style={font=\scriptsize}
 ]
  \addplot[mark=square,
           purple,
           ] coordinates {
            (0.56, 31.0)
            (1.1, 88.5)
            (1.7, 110.5)
            (3, 122.0)
            (7.1, 126.5)};
    \addplot[mark=triangle, mark size=0.7ex,
           orange,
           ] coordinates {
            (0.56, 81.0)
            (1.1, 100.0)
            (1.7, 104.0)
            (3, 114.0)
            (7.1, 112.0)};
    \addplot[mark=asterisk, mark size=0.8ex, teal,
           ] coordinates {
            (0.56, 83.0)
            (1.1, 105.0)
            (1.7, 113.0)
            (3, 117.0)
            (7.1, 123.0)};
    \addplot[mark=o, blue!60,
           ] coordinates {
            (0.56, 53.5)
            (1.1, 92.5)
            (1.7, 97.0)
            (3, 92.0)
            (7.1, 121.0)};
\end{axis}
\end{tikzpicture}

%% file: plot_size_fft_bloom_on_fr.tex
\begin{tikzpicture}
\begin{axis}[
 xlabel=French,
 width=1.15\linewidth,height=\linewidth,
 ymax=150, ymin=0,
 xtick={0.56, 1.1, 1.7, 3, 7.1},
 xmode=log, log basis x={2},
 xticklabel=\pgfmathparse{2^\tick}\pgfmathprintnumber{\pgfmathresult},
 yticklabel style={
        /pgf/number format/fixed,
        /pgf/number format/fixed zerofill,
        /pgf/number format/precision=0
 },
 tick label style={font=\tiny},
 ymajorgrids=true,
 grid style=dashed,
 xlabel style={font=\scriptsize}
 ]
  \addplot[mark=square,
           purple,
           ] coordinates {
            (0.56, 41.5)
            (1.1, 100.0)
            (1.7, 120.5)
            (3, 119.0)
            (7.1, 134.0)};
    \addplot[mark=triangle, mark size=0.7ex,
           orange,
           ] coordinates {
            (0.56, 88.5)
            (1.1, 99.0)
            (1.7, 112.5)
            (3, 116.0)
            (7.1, 109.0)};
    \addplot[mark=asterisk, mark size=0.8ex, teal,
           ] coordinates {
            (0.56, 76.0)
            (1.1, 99.5)
            (1.7, 110.5)
            (3, 115.0)
            (7.1, 125.0)};
    \addplot[mark=o, blue!60,
           ] coordinates {
            (0.56, 53.0)
            (1.1, 85.0)
            (1.7, 104.0)
            (3, 90.5)
            (7.1, 119.5)};
\end{axis}
\end{tikzpicture}

%% file: plot_size_fft_bloom_on_bg.tex
\begin{tikzpicture}
\begin{axis}[
 xlabel=Bulgarian,
 width=1.15\linewidth,height=\linewidth,
 ymax=150, ymin=0,
 xtick={0.56, 1.1, 1.7, 3, 7.1}, 
 xmode=log, log basis x={2},
 xticklabel=\pgfmathparse{2^\tick}\pgfmathprintnumber{\pgfmathresult},
 yticklabel style={
        /pgf/number format/fixed,
        /pgf/number format/fixed zerofill,
        /pgf/number format/precision=0
 },
 tick label style={font=\tiny},
 ymajorgrids=true,
 grid style=dashed,
 xlabel style={font=\scriptsize}
 ]
  \addplot[mark=square,
           purple,
           ] coordinates {
            (0.56, 42.0)
            (1.1, 69.0)
            (1.7, 81.0)
            (3, 105.5)
            (7.1, 63.5)};
    \addplot[mark=triangle, mark size=0.7ex,
           orange,
           ] coordinates {
            (0.56, 54.0)
            (1.1, 70.0)
            (1.7, 81.5)
            (3, 101.5)
            (7.1, 90.0)};
    \addplot[mark=asterisk, mark size=0.8ex, teal,
           ] coordinates {
            (0.56, 74.5)
            (1.1, 81.5)
            (1.7, 75.0)
            (3, 101.0)
            (7.1, 99.0)};
    \addplot[mark=o, blue!60,
           ] coordinates {
            (0.56, 0)
            (1.1, 4.5)
            (1.7, 0)
            (3, 3.0)
            (7.1, 12)};
\end{axis}
\end{tikzpicture}

%% file: plot_size_fft_bloom_on_ru.tex
\begin{tikzpicture}
\begin{axis}[
 xlabel=Russian,
 width=1.15\linewidth,height=\linewidth,
 ymax=150, ymin=0,
 xtick={0.56, 1.1, 1.7, 3, 7.1},
 xmode=log, log basis x={2},
 xticklabel=\pgfmathparse{2^\tick}\pgfmathprintnumber{\pgfmathresult},
 yticklabel style={
        /pgf/number format/fixed,
        /pgf/number format/fixed zerofill,
        /pgf/number format/precision=0
 },
 tick label style={font=\tiny},
 ymajorgrids=true,
 grid style=dashed,
 xlabel style={font=\scriptsize}
 ]
  \addplot[mark=square,
           purple,
           ] coordinates {
            (0.56, 39.0)
            (1.1, 76.5)
            (1.7, 94.0)
            (3, 103.0)
            (7.1, 106.5)};
    \addplot[mark=triangle, mark size=0.7ex,
           orange,
           ] coordinates {
            (0.56, 44.0)
            (1.1, 65.0)
            (1.7, 75.5)
            (3, 102.5)
            (7.1, 101.0)};
    \addplot[mark=asterisk, mark size=0.8ex, teal,
           ] coordinates {
            (0.56, 73.0)
            (1.1, 51.0)
            (1.7, 92.0)
            (3, 102.5)
            (7.1, 112.0)};
    \addplot[mark=o, blue!60,
           ] coordinates {
            (0.56, 3.0)
            (1.1, 12.0)
            (1.7, 7.0)
            (3, 6.0)
            (7.1, 36.0)};
\end{axis}
\end{tikzpicture}

%% file: plot_size_fft_bloom_on_zh.tex
\begin{tikzpicture}
\begin{axis}[
 xlabel=Chinese,
 width=1.15\linewidth,height=\linewidth,
 ymax=150, ymin=0,
 xtick={0.56, 1.1, 1.7, 3, 7.1},
 xmode=log, log basis x={2},
 xticklabel=\pgfmathparse{2^\tick}\pgfmathprintnumber{\pgfmathresult},
 yticklabel style={
        /pgf/number format/fixed,
        /pgf/number format/fixed zerofill,
        /pgf/number format/precision=0
 },
 tick label style={font=\tiny},
 ymajorgrids=true,
 grid style=dashed,
 xlabel style={font=\scriptsize}
 ]
  \addplot[mark=square,
           purple,
           ] coordinates {
            (0.56, 41.0)
            (1.1, 90.5)
            (1.7, 94.5)
            (3, 116.5)
            (7.1, 103.0)};
    \addplot[mark=triangle, mark size=0.7ex,
           orange,
           ] coordinates {
            (0.56, 47.5)
            (1.1, 70.5)
            (1.7, 87.0)
            (3, 110.5)
            (7.1, 102.0)};
    \addplot[mark=asterisk, mark size=0.8ex, teal,
           ] coordinates {
            (0.56, 32.0)
            (1.1, 78.5)
            (1.7, 101.0)
            (3, 110.0)
            (7.1, 115.5)};
    \addplot[mark=o, blue!60,
           ] coordinates {
            (0.56, 29.0)
            (1.1, 65.0)
            (1.7, 72.5)
            (3, 90.5)
            (7.1, 110.0)};
\end{axis}
\end{tikzpicture}

%% file: plot_size_lora_bloom_on_bn.tex
\begin{tikzpicture}
\begin{axis}[
 xlabel={LoRA, Bengali},
 width=1.15\linewidth,height=\linewidth,
 ymax=120, ymin=-0.01,
 xtick={0.56, 1.1, 1.7, 3, 7.1},
 xmode=log, log basis x={2},
 xticklabel=\pgfmathparse{2^\tick}\pgfmathprintnumber{\pgfmathresult},
 yticklabel style={
        /pgf/number format/fixed,
        /pgf/number format/fixed zerofill,
        /pgf/number format/precision=0
 },
 tick label style={font=\tiny},
 ymajorgrids=true,
 grid style=dashed,
 xlabel style={font=\scriptsize}
 ]
  \addplot[mark=square,
           purple,
           ] coordinates {
            (0.56, 52)
            (1.1, 62)
            (1.7, 68)
            (3, 88.5)
            (7.1, 107)};
    \addplot[mark=triangle, mark size=0.7ex,
           orange,
           ] coordinates {
            (0.56, 67)
            (1.1, 73)
            (1.7, 87)
            (3, 97.5)
            (7.1, 105.5)};
    \addplot[mark=o,
             blue!60,
           ] coordinates {
            (0.56, 22)
            (1.1, 79.5)
            (1.7, 36.0)
            (3, 32.0)
            (7.1, 54)};
\end{axis}
\end{tikzpicture}

%% file: plot_size_fft_bloom_on_bn.tex
\begin{tikzpicture}
\begin{axis}[
 xlabel={Full, Bengali},
 width=1.15\linewidth,height=\linewidth,
 ymax=120, ymin=-0.01,
 xtick={0.56, 1.1, 1.7, 3, 7.1},
 xmode=log, log basis x={2},
 xticklabel=\pgfmathparse{2^\tick}\pgfmathprintnumber{\pgfmathresult},
 yticklabel style={
        /pgf/number format/fixed,
        /pgf/number format/fixed zerofill,
        /pgf/number format/precision=0
 },
 tick label style={font=\tiny},
 ymajorgrids=true,
 grid style=dashed,
 xlabel style={font=\scriptsize}
 ]
  \addplot[mark=square,
           purple,
           ] coordinates {
            (0.56, 0.0)
            (1.1, 0.0)
            (1.7, 49.5)
            (3, 21.0)
            (7.1, 96.5)};
    \addplot[mark=triangle, mark size=0.7ex,
           orange,
           ] coordinates {
            (0.56, 9.5)
            (1.1, 0.0)
            (1.7, 81.0)
            (3, 46.0)
            (7.1, 111.0)};
    \addplot[mark=o, blue!60,
           ] coordinates {
            (0.56, 18.0)
            (1.1, 1.0)
            (1.7, 85.5)
            (3, 52.0)
            (7.1, 83.0)
            };
\end{axis}
\end{tikzpicture}

%% file: plot_size_lora_bloom_on_no.tex
\begin{tikzpicture}
\begin{axis}[
 xlabel={LoRA, Norwegian},
 width=1.15\linewidth,height=\linewidth,
 ymax=50, ymin=-0.01,
 xtick={0.56, 1.1, 1.7, 3, 7.1},
 xmode=log, log basis x={2},
 xticklabel=\pgfmathparse{2^\tick}\pgfmathprintnumber{\pgfmathresult},
 yticklabel style={
        /pgf/number format/fixed,
        /pgf/number format/fixed zerofill,
        /pgf/number format/precision=0
 },
 tick label style={font=\tiny},
 ymajorgrids=true,
 grid style=dashed,
 xlabel style={font=\scriptsize}
 ]
  \addplot[mark=square,
           purple,
           ] coordinates {
            (0.56, 33)
            (1.1, 33)
            (1.7, 35)
            (3, 32)
            (7.1, 24)};
    \addplot[mark=triangle, mark size=0.7ex,
           orange,
           ] coordinates {
            (0.56, 39)
            (1.1, 35)
            (1.7, 46)
            (3, 39)
            (7.1, 44)};
    \addplot[mark=o,
             blue!60,
           ] coordinates {
            (0.56, 3)
            (1.1, 19)
            (1.7, 2)
            (3, 0)
            (7.1, 1)};
\end{axis}
\end{tikzpicture}

%% file: plot_size_fft_bloom_on_no.tex
\begin{tikzpicture}
\begin{axis}[
 xlabel={Full, Norwegian},
 width=1.15\linewidth,height=\linewidth,
 ymax=50, ymin=-0.01,
 xtick={0.56, 1.1, 1.7, 3, 7.1},
 xmode=log, log basis x={2},
 xticklabel=\pgfmathparse{2^\tick}\pgfmathprintnumber{\pgfmathresult},
 yticklabel style={
        /pgf/number format/fixed,
        /pgf/number format/fixed zerofill,
        /pgf/number format/precision=0
 },
 tick label style={font=\tiny},
 ymajorgrids=true,
 grid style=dashed,
 xlabel style={font=\scriptsize}
 ]
  \addplot[mark=square,
           purple,
           ] coordinates {
            (0.56, 12.0)
            (1.1, 2.0)
            (1.7, 7.0)
            (3, 1.0)
            (7.1, 3.0)};
    \addplot[mark=triangle, mark size=0.7ex,
           orange,
           ] coordinates {
            (0.56, 9.0)
            (1.1, 4.0)
            (1.7, 9.0)
            (3, 1.0)
            (7.1, 6.0)};
    \addplot[mark=o, blue!60,
           ] coordinates {
            (0.56, 15.0)
            (1.1, 16.0)
            (1.7, 7.0)
            (3, 8.0)
            (7.1, 2.0)
            };
\end{axis}
\end{tikzpicture}

%% file: plot_robustness_lora_bloom.tex
\begin{tikzpicture}
\begin{axis}[
 xlabel={LoRA, BLOOM},
 width=1.15\linewidth,height=\linewidth,
 ymax=128, ymin=1,
 xtick={0.56, 1.1, 1.7, 3, 7.1}, 
 xmode=log, log basis x={2},
 ymode=log, log basis y={2},
 xticklabel=\pgfmathparse{2^\tick}\pgfmathprintnumber{\pgfmathresult},
 yticklabel=\pgfmathparse{2^\tick}\pgfmathprintnumber{\pgfmathresult},
 yticklabel style={
        /pgf/number format/fixed,
        /pgf/number format/fixed zerofill,
        /pgf/number format/precision=0
 },
 tick label style={font=\tiny},
 ymajorgrids=true,
 grid style=dashed,
 xlabel style={font=\scriptsize}
 ]
  \addplot[
           dashed,
           mark options=solid,mark=square,
           purple,
           ] coordinates {
            (0.56, 2.3)
            (1.1, 2.0)
            (1.7, 3.7)
            (3, 3.2)
            (7.1, 3.7)};
    \addplot[
           dashed,
           mark options=solid,mark=triangle, mark size=0.7ex,
           orange,
           ] coordinates {
            (0.56, 3.8)
            (1.1, 3.0)
            (1.7, 3.7)
            (3, 3.9)
            (7.1, 4.8)};
    \addplot[
           dashed,
           mark options=solid,mark=asterisk, mark size=0.8ex, teal,
           ] coordinates {
            (0.56, 2.0)
            (1.1, 2.7)
            (1.7, 2.7)
            (3, 4.0)
            (7.1, 4.0)};
    \addplot[
           dashed,
           mark options=solid,mark=o, blue!60,
           ] coordinates {
            (0.56, 51.3)
            (1.1, 10.4)
            (1.7, 44.2)
            (3, 61.6)
            (7.1, 58.7)};
\end{axis}
\end{tikzpicture}

%% file: plot_robustness_lora_pythia.tex
\begin{tikzpicture}
\begin{axis}[
 xlabel={LoRA, Pythia},
 width=1.15\linewidth,height=\linewidth,
 ymax=128, ymin=1,
 xtick={0.16, 0.41, 1.0, 2.8, 6.9, 12},
 xmode=log, log basis x={2},
 ymode=log, log basis y={2},
 xticklabel=\pgfmathparse{2^\tick}\pgfmathprintnumber{\pgfmathresult},
 yticklabel=\pgfmathparse{2^\tick}\pgfmathprintnumber{\pgfmathresult},
 yticklabel style={
        /pgf/number format/fixed,
        /pgf/number format/fixed zerofill,
        /pgf/number format/precision=0
 },
 tick label style={font=\tiny},
 ymajorgrids=true,
 grid style=dashed,
 xlabel style={font=\scriptsize}
 ]
  \addplot[
           dashed,
           mark options=solid,mark=square,
           purple,
           ] coordinates {
            (0.07, 20.5) 
            (0.16, 8.2)
            (0.41, 1.7)
            (1.0, 1.7)
            (1.4, 3.3)
            (2.8, 2.5)
            (6.9, 1.5)
            (12, 2.2)};
    \addplot[
           dashed,
           mark options=solid,mark=triangle, mark size=0.7ex,
           orange,
           ] coordinates {
            (0.07, 20.5) 
            (0.16, 8.0)
            (0.41, 3.7)
            (1.0, 3.8)
            (1.4, 3.0)
            (2.8, 3.7)
            (6.9, 3.0)
            (12, 2.3)};
    \addplot[
           dashed,
           mark options=solid,mark=asterisk, mark size=0.8ex,
           teal,
           ] coordinates {
            (0.07, 20.8) 
            (0.16, 8.0)
            (0.41, 2.7)
            (1.0, 2.8)
            (1.4, 3.3)
            (2.8, 3.2)
            (6.9, 3.2)
            (12, 3.3)};
    \addplot[
           dashed,
           mark options=solid,mark=o,
             blue!60,
           ] coordinates {
            (0.07, 21.2) 
            (0.16, 7.1)
            (0.41, 20.3)
            (1.0, 41.2)
            (1.4, 57.7)
            (2.8, 38.8)
            (6.9, 42.4)
            (12, 33.4)};
\end{axis}
\end{tikzpicture}

%% file: plot_robustness_fft_bloom.tex
\begin{tikzpicture}
\begin{axis}[
 xlabel={Full, BLOOM},
 width=1.15\linewidth,height=\linewidth,
 ymax=128, ymin=1,
 xtick={0.56, 1.1, 1.7, 3, 7.1}, 
 xmode=log, log basis x={2},
 ymode=log, log basis y={2},
 xticklabel=\pgfmathparse{2^\tick}\pgfmathprintnumber{\pgfmathresult},
 yticklabel=\pgfmathparse{2^\tick}\pgfmathprintnumber{\pgfmathresult},
 yticklabel style={
        /pgf/number format/fixed,
        /pgf/number format/fixed zerofill,
        /pgf/number format/precision=0
 },
 tick label style={font=\scriptsize},
 ymajorgrids=true,
 grid style=dashed,
 xlabel style={font=\scriptsize}
 ]
  \addplot[
           dashed,
           mark options=solid,
           mark=square,
           purple,
           ] coordinates {
            (0.56, 18.2)
            (1.1, 10.8)
            (1.7, 7.1)
            (3, 4.5)
            (7.1, 9.0)};
    \addplot[
           dashed,
           mark options=solid,mark=triangle, mark size=0.7ex,
           orange,
           ] coordinates {
            (0.56, 15.4)
            (1.1, 9.8)
            (1.7, 8.7)
            (3, 5.8)
            (7.1, 8.2)};
    \addplot[
           dashed,
           mark options=solid,mark=asterisk, mark size=0.8ex, teal,
           ] coordinates {
            (0.56, 9.2)
            (1.1, 7.5)
            (1.7, 6.8)
            (3, 4.8)
            (7.1, 6.2)};
    \addplot[
           dashed,
           mark options=solid,mark=o, blue!60,
           ] coordinates {
            (0.56, 30.3)
            (1.1, 24.9)
            (1.7, 29.8)
            (3, 41.8)
            (7.1, 27.3)};
\end{axis}
\end{tikzpicture}

%% file: plot_robustness_lora_fft_bloom_pythia_legend.tex
\begin{tikzpicture}
\begin{customlegend}[
    legend columns=1,
    legend style={
            column sep=3ex,
            font=\scriptsize,
    },
    legend entries={
            \hspace{-2.5ex}multilingual,
            \hspace{-2.5ex}monolingual,
            \hspace{-2.5ex}multilingual-downsample,
            \hspace{-2.5ex}English model,
    }
]
\addlegendimage{
    dashed,
    mark options=solid,
    mark=square,
    purple,
}
\addlegendimage{
    dashed,
    mark options=solid,
    mark=asterisk,
    mark size=0.8ex,
    teal,
}
\addlegendimage{
    dashed,
    mark options=solid,
    mark=triangle,
    mark size=0.7ex,
    orange,
}
\addlegendimage{
    dashed,
    mark options=solid,
    mark=o,
    blue!60,
}
\end{customlegend}
\end{tikzpicture}

%% file: plot_family_lora_7b_on_en.tex
\begin{tikzpicture}
\begin{axis}[
    xlabel=English,
    width=1.1\linewidth,height=0.9\linewidth,
    ybar,
    ymax=140,ymin=80,
    bar width=0.85ex,
    enlarge y limits={0.4},
    enlarge x limits=0.12,
    symbolic x coords={Baichuan-2, BLOOM, LLaMA, OpenLLaMA, Pythia},
    xtick=data,
    xtick align=inside,
    xticklabel style={rotate=15, anchor=north, xshift=-0.2ex, yshift=-0.1ex},
    nodes near coords,
    nodes near coords style={font=\tiny, rotate=90, anchor=west},
    tick label style={font=\tiny},
 xlabel style={font=\scriptsize}
    ]
    \addplot [
        draw=black!75,
        fill=purple!30,
        text=black
    ]
    coordinates {
        (Baichuan-2, 126.0)
        (BLOOM, 113.0)
        (LLaMA, 138.0)
        (OpenLLaMA, 133.0)
        (Pythia, 120.5)
    };
    \addplot [
        draw=black!75,
        fill=orange!30,
        text=black
    ] coordinates {
        (Baichuan-2, 128.5)
        (BLOOM, 109.5)
        (LLaMA, 129.5)
        (OpenLLaMA, 126.5)
        (Pythia, 119.0)
    };
    \addplot [
        draw=black!75,
        fill=teal!30,
        text=black
    ] coordinates {
        (Baichuan-2, 126.5)
        (BLOOM, 122.0)
        (LLaMA, 133.5)
        (OpenLLaMA, 122)
        (Pythia, 115.0)
    };
\end{axis}
\end{tikzpicture}

%% file: plot_family_lora_7b_on_es.tex
\begin{tikzpicture}
\begin{axis}[
    xlabel=Spanish,
    width=1.1\linewidth,height=0.9\linewidth,
    ybar,
    ymax=140,ymin=80,
    bar width=0.85ex,
    enlarge y limits={0.4},
    enlarge x limits=0.12,
    symbolic x coords={Baichuan-2, BLOOM, LLaMA, OpenLLaMA, Pythia},
    xtick=data,
    xtick align=inside,
    xticklabel style={rotate=15, anchor=north, xshift=-0.2ex, yshift=-0.1ex},
    nodes near coords,
    nodes near coords style={font=\tiny, rotate=90, anchor=west},
    tick label style={font=\tiny},
 xlabel style={font=\scriptsize}
    ]
    \addplot [
        draw=black!75,
        fill=purple!30,
        text=black
    ]
    coordinates {
        (Baichuan-2, 127.5)
        (BLOOM, 122.0)
        (LLaMA, 140.5)
        (OpenLLaMA, 122.0)
        (Pythia, 119.0)
    };
    \addplot [
        draw=black!75,
        fill=orange!30,
        text=black
    ] coordinates {
        (Baichuan-2, 124.0)
        (BLOOM, 115.0)
        (LLaMA, 131.5)
        (OpenLLaMA, 122.5)
        (Pythia, 118.5)
    };
    \addplot [
        draw=black!75,
        fill=teal!30,
        text=black
    ] coordinates {
        (Baichuan-2, 121.0)
        (BLOOM, 116.5)
        (LLaMA, 127.0)
        (OpenLLaMA, 113.5)
        (Pythia, 100.5)
    };
\end{axis}
\end{tikzpicture}

%% file: plot_family_lora_7b_on_fr.tex
\begin{tikzpicture}
\begin{axis}[
    xlabel=French,
    width=1.1\linewidth,height=0.9\linewidth,
    ybar,
    ymax=140,ymin=80,
    bar width=0.85ex,
    enlarge y limits={0.4},
    enlarge x limits=0.12,
    symbolic x coords={Baichuan-2, BLOOM, LLaMA, OpenLLaMA, Pythia},
    xtick=data,
    xtick align=inside,
    xticklabel style={rotate=15, anchor=north, xshift=-0.2ex, yshift=-0.1ex},
    nodes near coords,
    nodes near coords style={font=\tiny, rotate=90, anchor=west},
    tick label style={font=\tiny},
 xlabel style={font=\scriptsize}
    ]
    \addplot [
        draw=black!75,
        fill=purple!30,
        text=black
    ]
    coordinates {
        (Baichuan-2, 121.5)
        (BLOOM, 125.5)
        (LLaMA, 137.0)
        (OpenLLaMA, 119.0)
        (Pythia, 119.0)
    };
    \addplot [
        draw=black!75,
        fill=orange!30,
        text=black
    ] coordinates {
        (Baichuan-2, 125.0)
        (BLOOM, 116.0)
        (LLaMA, 127.5)
        (OpenLLaMA, 128.0)
        (Pythia, 106.0)
    };
    \addplot [
        draw=black!75,
        fill=teal!30,
        text=black
    ] coordinates {
        (Baichuan-2, 114.0)
        (BLOOM, 104.5)
        (LLaMA, 134.0)
        (OpenLLaMA, 117.0)
        (Pythia, 96.0)
    };
\end{axis}
\end{tikzpicture}

%% file: plot_family_lora_7b_on_bg.tex
\begin{tikzpicture}
\begin{axis}[
    xlabel=Bulgarian,
    width=1.1\linewidth,height=0.9\linewidth,
    ybar,
    ymax=140,ymin=80,
    bar width=0.85ex,
    enlarge y limits={0.4},
    enlarge x limits=0.12,
    symbolic x coords={Baichuan-2, BLOOM, LLaMA, OpenLLaMA, Pythia},
    xtick=data,
    xtick align=inside,
    xticklabel style={rotate=15, anchor=north, xshift=-0.2ex, yshift=-0.1ex},
    nodes near coords,
    nodes near coords style={font=\tiny, rotate=90, anchor=west},
    tick label style={font=\tiny},
 xlabel style={font=\scriptsize}
    ]
    \addplot [
        draw=black!75,
        fill=purple!30,
        text=black
    ]
    coordinates {
        (Baichuan-2, 122.0)
        (BLOOM, 90.5)
        (LLaMA, 119.5)
        (OpenLLaMA, 110.0)
        (Pythia, 99.5)
    };
    \addplot [
        draw=black!75,
        fill=orange!30,
        text=black
    ] coordinates {
        (Baichuan-2, 118.5)
        (BLOOM, 75.5)
        (LLaMA, 111.5)
        (OpenLLaMA, 106.5)
        (Pythia, 102.5)
    };
    \addplot [
        draw=black!75,
        fill=teal!30,
        text=black
    ] coordinates {
        (Baichuan-2, 123.5)
        (BLOOM, 79.5)
        (LLaMA, 120.5)
        (OpenLLaMA, 105.5)
        (Pythia, 87.0)
    };
\end{axis}
\end{tikzpicture}

%% file: plot_family_lora_7b_on_ru.tex
\begin{tikzpicture}
\begin{axis}[
    xlabel=Russian,
    width=1.1\linewidth,height=0.9\linewidth,
    ybar,
    ymax=140,ymin=80,
    bar width=0.85ex,
    enlarge y limits={0.4},
    enlarge x limits=0.12,
    symbolic x coords={Baichuan-2, BLOOM, LLaMA, OpenLLaMA, Pythia},
    xtick=data,
    xtick align=inside,
    xticklabel style={rotate=15, anchor=north, xshift=-0.2ex, yshift=-0.1ex},
    nodes near coords,
    nodes near coords style={font=\tiny, rotate=90, anchor=west},
    tick label style={font=\tiny},
 xlabel style={font=\scriptsize}
    ]
    \addplot [
        draw=black!75,
        fill=purple!30,
        text=black
    ]
    coordinates {
        (Baichuan-2, 119.0)
        (BLOOM, 93.5)
        (LLaMA, 129.0)
        (OpenLLaMA, 115.0)
        (Pythia, 105.5)
    };
    \addplot [
        draw=black!75,
        fill=orange!30,
        text=black
    ] coordinates {
        (Baichuan-2, 123.0)
        (BLOOM, 81.0)
        (LLaMA, 116.0)
        (OpenLLaMA, 114.0)
        (Pythia, 107.0)
    };
    \addplot [
        draw=black!75,
        fill=teal!30,
        text=black
    ] coordinates {
        (Baichuan-2, 113.0)
        (BLOOM, 94.0)
        (LLaMA, 138.0)
        (OpenLLaMA, 103.0)
        (Pythia, 86.0)
    };
\end{axis}
\end{tikzpicture}

%% file: plot_family_lora_7b_on_zh.tex
\begin{tikzpicture}
\begin{axis}[
    xlabel=Chinese,
    width=1.1\linewidth,height=0.9\linewidth,
    ybar,
    ymax=140,ymin=80,
    bar width=0.85ex,
    enlarge y limits={0.4},
    enlarge x limits=0.12,
    symbolic x coords={Baichuan-2, BLOOM, LLaMA, OpenLLaMA, Pythia},
    xtick=data,
    xtick align=inside,
    xticklabel style={rotate=15, anchor=north, xshift=-0.2ex, yshift=-0.1ex},
    nodes near coords,
    nodes near coords style={font=\tiny, rotate=90, anchor=west},
    tick label style={font=\tiny},
 xlabel style={font=\scriptsize}
    ]
    \addplot [
        draw=black!75,
        fill=purple!30,
        text=black
    ]
    coordinates {
        (Baichuan-2, 125.0)
        (BLOOM, 119.5)
        (LLaMA, 95.0)
        (OpenLLaMA, 80.0)
        (Pythia, 98.5)
    };
    \addplot [
        draw=black!75,
        fill=orange!30,
        text=black
    ] coordinates {
        (Baichuan-2, 131.0)
        (BLOOM, 117.0)
        (LLaMA, 85.0)
        (OpenLLaMA, 75.0)
        (Pythia, 91.0)
    };
    \addplot [
        draw=black!75,
        fill=teal!30,
        text=black
    ] coordinates {
        (Baichuan-2, 132.0)
        (BLOOM, 105.0)
        (LLaMA, 118.5)
        (OpenLLaMA, 79.5)
        (Pythia, 80.0)
    };
\end{axis}
\end{tikzpicture}

%% file: plot_family_lora_7b_legend.tex
\begin{tikzpicture}
\begin{customlegend}[
    legend columns=-1,
    legend style={
            column sep=3ex,
            font=\scriptsize,
    },
    legend entries={
            \hspace{-2.5ex}multilingual,
            \hspace{-2.5ex}multilingual-downsample,
            \hspace{-2.5ex}monolingual,
    }
]
\addlegendimage{
        area legend,
        draw=black!75,
        fill=purple!30,
}
\addlegendimage{
        area legend,
        draw=black!75,
        fill=orange!30,
}
\addlegendimage{
        area legend,
        draw=black!75,
        fill=teal!30,
}
\end{customlegend}
\end{tikzpicture}

%% file: table_hyperparameters.tex
\begin{table}[ht]
\centering\small
\begin{tabular}{lll}
\toprule
\multicolumn{1}{c}{Method} & \multicolumn{1}{c}{Hyperparameter} & \multicolumn{1}{c}{Value} \\
\midrule
\multirow{7}{*}{LoRA} & LoRA modules                  & query, key, value \\
& \phantom{LoRA} rank           & 8 \\
& \phantom{LoRA} scaling factor & 16 \\
& \phantom{LoRA} dropout        & 0.05 \\
& learning rate                 & 3e\textsuperscript{-4} \\
& global batch size             & 128 \\
& epochs                        & 5 \\
\cdashlinelr{1-3}
\multirow{3}{*}{FFT} & learning rate & 2e\textsuperscript{-5} \\
& global batch size & 256 \\
& epochs & 3 \\
\bottomrule
\end{tabular}
\caption{Hyperparameter configurations of LoRA and full-parameter fine-tuning}
\label{tab:hyperparameters}
\end{table}

%% file: plot_size_lora_pythia_on_en.tex
\begin{tikzpicture}
\begin{axis}[
 xlabel=English,
 width=1.15\linewidth,height=\linewidth,
 ymax=150, ymin=0,
 xtick={0.16, 0.41, 1.0, 2.8, 6.9, 12},
 xmode=log, log basis x={2},
 xticklabel=\pgfmathparse{2^\tick}\pgfmathprintnumber{\pgfmathresult},
 yticklabel style={
        /pgf/number format/fixed,
        /pgf/number format/fixed zerofill,
        /pgf/number format/precision=0
 },
 tick label style={font=\tiny},
 ymajorgrids=true,
 grid style=dashed,
 xlabel style={font=\scriptsize}
 ]
  \addplot[mark=square,
           purple,
           ] coordinates {
            (0.07, 45.5) 
            (0.16, 58.0)
            (0.41, 90.0)
            (1.0, 104.0)
            (1.4, 108.5)
            (2.8, 121.0)
            (6.9, 120.5)
            (12, 129.0)};
    \addplot[mark=triangle, mark size=0.7ex,
           orange,
           ] coordinates {
            (0.07, 45.0) 
            (0.16, 56.0)
            (0.41, 81.0)
            (1.0, 107.5)
            (1.4, 103.5)
            (2.8, 114.5)
            (6.9, 119.0)
            (12, 120.0)};
    \addplot[mark=asterisk, mark size=0.8ex,
           teal,
           ] coordinates {
            (0.07, 43.0) 
            (0.16, 59.5)
            (0.41, 84.0)
            (1.0, 100.5)
            (1.4, 110.0)
            (2.8, 109.0)
            (6.9, 115.0)
            (12, 126.0)};
\end{axis}
\end{tikzpicture}

%% file: plot_size_lora_pythia_on_es.tex
\begin{tikzpicture}
\begin{axis}[
 xlabel=Spanish,
 width=1.15\linewidth,height=\linewidth,
 ymax=150, ymin=0,
 xtick={0.16, 0.41, 1.0, 2.8, 6.9, 12},
 xmode=log, log basis x={2},
 xticklabel=\pgfmathparse{2^\tick}\pgfmathprintnumber{\pgfmathresult},
 yticklabel style={
        /pgf/number format/fixed,
        /pgf/number format/fixed zerofill,
        /pgf/number format/precision=0
 },
 tick label style={font=\tiny},
 ymajorgrids=true,
 grid style=dashed,
 xlabel style={font=\scriptsize}
 ]
  \addplot[mark=square,
           purple,
           ] coordinates {
            (0.07, 17.0) 
            (0.16, 48.0)
            (0.41, 66.5)
            (1.0, 87.5)
            (1.4, 86.5)
            (2.8, 98.5)
            (6.9, 119.0)
            (12, 120.5)};
    \addplot[mark=triangle, mark size=0.7ex,
           orange,
           ] coordinates {
            (0.07, 17.0) 
            (0.16, 48.0)
            (0.41, 61.0)
            (1.0, 72.5)
            (1.4, 88.5)
            (2.8, 108.5)
            (6.9, 118.5)
            (12, 126.5)};
    \addplot[mark=asterisk, mark size=0.8ex,
           teal,
           ] coordinates {
            (0.07, 17.0) 
            (0.16, 48.0)
            (0.41, 58.0)
            (1.0, 67.0)
            (1.4, 81.5)
            (2.8, 98.5)
            (6.9, 100.5)
            (12, 101.5)};
    \addplot[mark=o,
             blue!60,
           ] coordinates {
            (0.07, 17.0) 
            (0.16, 48.0)
            (0.41, 22.0)
            (1.0, 19.0)
            (1.4, 12.0)
            (2.8, 70.5)
            (6.9, 70.0)
            (12, 92.0)};
\end{axis}
\end{tikzpicture}

%% file: plot_size_lora_pythia_on_fr.tex
\begin{tikzpicture}
\begin{axis}[
 xlabel=French,
 width=1.15\linewidth,height=\linewidth,
 ymax=150, ymin=0,
 xtick={0.16, 0.41, 1.0, 2.8, 6.9, 12},
 xmode=log, log basis x={2},
 xticklabel=\pgfmathparse{2^\tick}\pgfmathprintnumber{\pgfmathresult},
 yticklabel style={
        /pgf/number format/fixed,
        /pgf/number format/fixed zerofill,
        /pgf/number format/precision=0
 },
 tick label style={font=\tiny},
 ymajorgrids=true,
 grid style=dashed,
 xlabel style={font=\scriptsize}
 ]
  \addplot[mark=square,
           purple,
           ] coordinates {
            (0.07, 25.0) 
            (0.16, 46.0)
            (0.41, 71.0)
            (1.0, 79.0)
            (1.4, 94.0)
            (2.8, 117.0)
            (6.9, 119.0)
            (12, 124.5)};
    \addplot[mark=triangle, mark size=0.7ex,
           orange,
           ] coordinates {
            (0.07, 25.0) 
            (0.16, 46.0)
            (0.41, 54.0)
            (1.0, 81.5)
            (1.4, 75.5)
            (2.8, 103.0)
            (6.9, 106.0)
            (12, 119.0)};
    \addplot[mark=asterisk, mark size=0.8ex,
           teal,
           ] coordinates {
            (0.07, 25.0) 
            (0.16, 45.0)
            (0.41, 57.0)
            (1.0, 68.0)
            (1.4, 83.5)
            (2.8, 94.5)
            (6.9, 96.0)
            (12, 106.5)};
    \addplot[mark=o,
             blue!60,
           ] coordinates {
            (0.07, 24.0) 
            (0.16, 43.0)
            (0.41, 10.0)
            (1.0, 9.0)
            (1.4, 10.0)
            (2.8, 50)
            (6.9, 87.5)
            (12, 84.0)};
\end{axis}
\end{tikzpicture}

%% file: plot_size_lora_pythia_on_bg.tex
\begin{tikzpicture}
\begin{axis}[
 xlabel=Bulgarian,
 width=1.15\linewidth,height=\linewidth,
 ymax=150, ymin=0,
 xtick={0.16, 0.41, 1.0, 2.8, 6.9, 12},
 xmode=log, log basis x={2},
 xticklabel=\pgfmathparse{2^\tick}\pgfmathprintnumber{\pgfmathresult},
 yticklabel style={
        /pgf/number format/fixed,
        /pgf/number format/fixed zerofill,
        /pgf/number format/precision=0
 },
 tick label style={font=\tiny},
 ymajorgrids=true,
 grid style=dashed,
 xlabel style={font=\scriptsize}
 ]
  \addplot[mark=square,
           purple,
           ] coordinates {
            (0.07, 30.0) 
            (0.16, 37.0)
            (0.41, 54.0)
            (1.0, 59)
            (1.4, 79.0)
            (2.8, 104.0)
            (6.9, 99.5)
            (12, 114.0)};
    \addplot[mark=triangle, mark size=0.7ex,
           orange,
           ] coordinates {
            (0.07, 30.0) 
            (0.16, 37.0)
            (0.41, 48.0)
            (1.0, 55.0)
            (1.4, 68.0)
            (2.8, 79.5)
            (6.9, 102.5)
            (12, 104.5)};
    \addplot[mark=asterisk, mark size=0.8ex,
           teal,
           ] coordinates {
            (0.07, 30.0) 
            (0.16, 37.0)
            (0.41, 50.0)
            (1.0, 61.0)
            (1.4, 66.0)
            (2.8, 73.5)
            (6.9, 87.0)
            (12, 97.5)};
    \addplot[mark=o,
             blue!60,
           ] coordinates {
            (0.07, 50.0) 
            (0.16, 37.0)
            (0.41, 46.0)
            (1.0, 42.0)
            (1.4, 9.0)
            (2.8, 36.5)
            (6.9, 44.0)
            (12, 42.0)};
\end{axis}
\end{tikzpicture}

%% file: plot_size_lora_pythia_on_ru.tex
\begin{tikzpicture}
\begin{axis}[
 xlabel=Russian,
 width=1.15\linewidth,height=\linewidth,
 ymax=150, ymin=0,
 xtick={0.16, 0.41, 1.0, 2.8, 6.9, 12},
 xmode=log, log basis x={2},
 xticklabel=\pgfmathparse{2^\tick}\pgfmathprintnumber{\pgfmathresult},
 yticklabel style={
        /pgf/number format/fixed,
        /pgf/number format/fixed zerofill,
        /pgf/number format/precision=0
 },
 tick label style={font=\tiny},
 ymajorgrids=true,
 grid style=dashed,
 xlabel style={font=\scriptsize}
 ]
  \addplot[mark=square,
           purple,
           ] coordinates {
            (0.07, 43.0) 
            (0.16, 44.0)
            (0.41, 60.0)
            (1.0, 69.0)
            (1.4, 78.0)
            (2.8, 102.5)
            (6.9, 105.5)
            (12, 110.5)};
    \addplot[mark=triangle, mark size=0.7ex,
           orange,
           ] coordinates {
            (0.07, 43.0) 
            (0.16, 44.0)
            (0.41, 52.0)
            (1.0, 61.5)
            (1.4, 78.0)
            (2.8, 89.5)
            (6.9, 107.0)
            (12, 109.5)};
    \addplot[mark=asterisk, mark size=0.8ex,
           teal,
           ] coordinates {
            (0.07, 43.0) 
            (0.16, 44.0)
            (0.41, 54.0)
            (1.0, 62.0)
            (1.4, 67.5)
            (2.8, 90.0)
            (6.9, 86.0)
            (12, 97.5)};
    \addplot[mark=o,
             blue!60,
           ] coordinates {
            (0.07, 43.0) 
            (0.16, 44.0)
            (0.41, 39.0)
            (1.0, 26.0)
            (1.4, 12.0)
            (2.8, 49.5)
            (6.9, 36.0)
            (12, 58.5)};
\end{axis}
\end{tikzpicture}

%% file: plot_size_lora_pythia_on_zh.tex
\begin{tikzpicture}
\begin{axis}[
 xlabel=Chinese,
 width=1.15\linewidth,height=\linewidth,
 ymax=150, ymin=0,
 xtick={0.16, 0.41, 1.0, 2.8, 6.9, 12},
 xmode=log, log basis x={2},
 xticklabel=\pgfmathparse{2^\tick}\pgfmathprintnumber{\pgfmathresult},
 yticklabel style={
        /pgf/number format/fixed,
        /pgf/number format/fixed zerofill,
        /pgf/number format/precision=0
 },
 tick label style={font=\tiny},
 ymajorgrids=true,
 grid style=dashed,
 xlabel style={font=\scriptsize}
 ]
  \addplot[mark=square,
           purple,
           ] coordinates {
            (0.07, 31.0) 
            (0.16, 36.0)
            (0.41, 56.0)
            (1.0, 82.5)
            (1.4, 83.5)
            (2.8, 91.0)
            (6.9, 98.5)
            (12, 104.0)};
    \addplot[mark=triangle, mark size=0.7ex,
           orange,
           ] coordinates {
            (0.07, 31.0) 
            (0.16, 38.0)
            (0.41, 51.0)
            (1.0, 62.5)
            (1.4, 64.0)
            (2.8, 79.0)
            (6.9, 91.0)
            (12, 98.0)};
    \addplot[mark=asterisk, mark size=0.8ex,
           teal,
           ] coordinates {
            (0.07, 31.0) 
            (0.16, 33.0)
            (0.41, 56.0)
            (1.0, 65.0)
            (1.4, 68.0)
            (2.8, 80.5)
            (6.9, 80.0)
            (12, 96.5)};
    \addplot[mark=o,
             blue!60,
           ] coordinates {
            (0.07, 33.0) 
            (0.16, 38.0)
            (0.41, 26.0)
            (1.0, 13.0)
            (1.4, 10.0)
            (2.8, 20.0)
            (6.9, 12.0)
            (12, 34.5)};
\end{axis}
\end{tikzpicture}

%% file: plot_size_lora_pythia_legend.tex
\begin{tikzpicture}
\begin{customlegend}[
    legend columns=2,
    legend style={
            column sep=3ex,
            font=\scriptsize,
    },
    legend entries={
            \hspace{-2.5ex}monolingual,
            \hspace{-2.5ex}multilingual,
            \hspace{-2.5ex}multilingual-downsample,
            \hspace{-2.5ex}English model,
    }
]
\addlegendimage{
    mark=asterisk,
    mark size=0.8ex,
    teal,
}
\addlegendimage{
    mark=square,
    purple,
}
\addlegendimage{
    mark=triangle,
    mark size=0.7ex,
    orange,
}
\addlegendimage{
    mark=o,
    blue!60,
}
\end{customlegend}
\end{tikzpicture}